%% file: iclr2027_conference.tex
\documentclass{article}
\ifdefined\pdfminorversion\pdfminorversion=7\fi
\usepackage{iclr2027_conference,times}
\input{math_commands.tex}

\usepackage[T1]{fontenc}
\usepackage[utf8]{inputenc}
\usepackage{silence}
\usepackage{microtype}
\usepackage{graphicx}
\usepackage{booktabs}
\usepackage{adjustbox}
\usepackage{multirow}
\usepackage{threeparttable}
\usepackage{tabularx}
\usepackage{makecell}
\usepackage{array}
\usepackage{amsmath,amssymb,bm}
\usepackage{mathtools}
\usepackage[table]{xcolor}
\usepackage{algorithm}
\usepackage{algpseudocode}
\usepackage{float}
\usepackage{wrapfig}
\usepackage{needspace}
\usepackage{placeins}
\usepackage{subcaption}
\usepackage{pifont}
\usepackage{hyperref}
\usepackage{url}
\hypersetup{colorlinks=true,citecolor=blue,linkcolor=red,urlcolor=blue}
\graphicspath{{figs/}}
\definecolor{fatarow}{HTML}{FFF5E0}
\definecolor{avgrow}{HTML}{FFF5E0}
\definecolor{lightgray}{RGB}{242,242,242}
\definecolor{verylightblue}{RGB}{239,246,252}
\definecolor{gaincolor}{RGB}{34,139,34}
\definecolor{headergray}{RGB}{238,241,245}
\definecolor{BudgetGray}{RGB}{224,224,224}
\definecolor{HeaderGray}{RGB}{242,242,242}
\newcolumntype{C}[1]{>{\centering\arraybackslash}p{#1}}
\newcolumntype{Y}{>{\centering\arraybackslash}X}

\newcommand{\SR}{\mathrm{SR}}
\newcommand{\CER}{\mathrm{CER}}
\newcommand{\CBR}{\mathrm{CBR}}
\newcommand{\Kprac}{K_{\mathrm{prac}}}
\newcommand{\Lattn}{\mathcal{L}_{\mathrm{attn}}}
\newcommand{\Lsem}{\mathcal{L}_{\mathrm{sem}}}
\newcommand{\Ltotal}{\mathcal{L}_{\mathrm{FATA}}}

\newcommand{\TopK}{\operatorname{TopK}}
\newcommand{\Clip}{\operatorname{clip}}
\newcommand{\Proj}{\Pi_{\|\delta\|_{\infty}\leq\epsilon}}
\newcommand{\stealthgain}[1]{\textcolor{gaincolor}{\scriptsize\ensuremath{\uparrow}#1\%}}

\title{Feature-Aware Token Attack for Compression-Triggered Stealthy Failures in Large Vision-Language Models}
\iclrfinalcopy

\author{%
\begin{tabular}{@{}l@{}}
\textbf{Shilinlu Yan\textsuperscript{1}}
\quad
\textbf{Bowen Chen\textsuperscript{1}}
\quad
\textbf{Yuechen Zhang\textsuperscript{2}}
\quad
\textbf{Zhenhong Zhou\textsuperscript{3}}
\quad
\textbf{Li Sun\textsuperscript{1}}
\quad
\textbf{Sen Su\textsuperscript{1,4}}
\end{tabular}
\\[2pt]
\textsuperscript{1}Beijing University of Posts and Telecommunications
\\
\textsuperscript{2}Jiangnan University
\\
\textsuperscript{3}Nanyang Technological University
\\
\textsuperscript{4}Chongqing University of Posts and Telecommunications
\\[2pt]
\texttt{\{lulu\_land,cbcbw,lsun,susen\}@bupt.edu.cn}
\\
\texttt{orange.zhangyc05@gmail.com}
\quad
\texttt{zhenhong001@e.ntu.edu.sg}
}

\begin{document}
\maketitle
\input{1-abstract}
\input{2-introduction}

\input{3-related_work}
\input{5-method}
\input{6-experiments}
\input{7-defense}
\input{8-conclution}
\bibliographystyle{iclr2027_conference}
\bibliography{custom}
\newpage
\appendix
\input{9-appendix}

\end{document}

%% file: math_commands.tex
\usepackage{amsmath,amsfonts,bm}

\def\eqref#1{equation~\ref{#1}}

\def\1{\bm{1}}

\DeclareMathAlphabet{\mathsfit}{\encodingdefault}{\sfdefault}{m}{sl}
\SetMathAlphabet{\mathsfit}{bold}{\encodingdefault}{\sfdefault}{bx}{n}



%% file: 1-abstract.tex
\begin{abstract}
Visual-token compression improves the efficiency of large vision-language models, but can expose failures that full-token evaluation misses. We study adversarial images that preserve full-token correctness yet induce errors after compression, even when both inference paths succeed on the clean image. Creating such failures is challenging because perturbing token importance can also damage the visual content needed for full-token inference. We propose \textbf{Feature-Aware Token Attack (FATA)}, which couples attention suppression with cosine-based feature preservation on a fixed set of salient clean-image tokens. In the primary LLaVA-1.5-7B setting, FATA uses only vision-encoder gradients, without access to the deployed compressor, token budget, or downstream task. Across four visually dependent task subsets and four compressors under a controlled reconstruction protocol, FATA achieves \(\SR=\mathbf{96.3\%}\) full-token accuracy retention and \(\CBR=\mathbf{22.1\%}\) conditional blinding, compared with \(89.8\%\) and \(15.7\%\) for CAA. Ablations support the role of both objectives in balancing compressed-path failure against full-token preservation. FATA also has the lowest measured detection rate among four attacks across three evaluated detectors at a \(5\%\) false-positive rate. These findings motivate assessing adversarial robustness jointly across full-token and compressed inference.
\end{abstract}

%% file: 2-introduction.tex
\section{Introduction}

Visual-token compression reduces the inference cost of large vision-language models (LVLMs) as image resolution and visual context grow \citep{an2026llavaov2,song2026evocomp}. Recent methods select informative tokens or merge redundant features while retaining much of the original task accuracy \citep{yang2025visionzip,zhang2025beyond,shang2025llava,tong2026flowcut}. Because these decisions determine which visual evidence remains available to the language model \citep{chen2024image}, their robustness to input perturbations is essential to reliable compressed inference.

\begin{figure}[t]
    \centering
    \includegraphics[width=\linewidth]{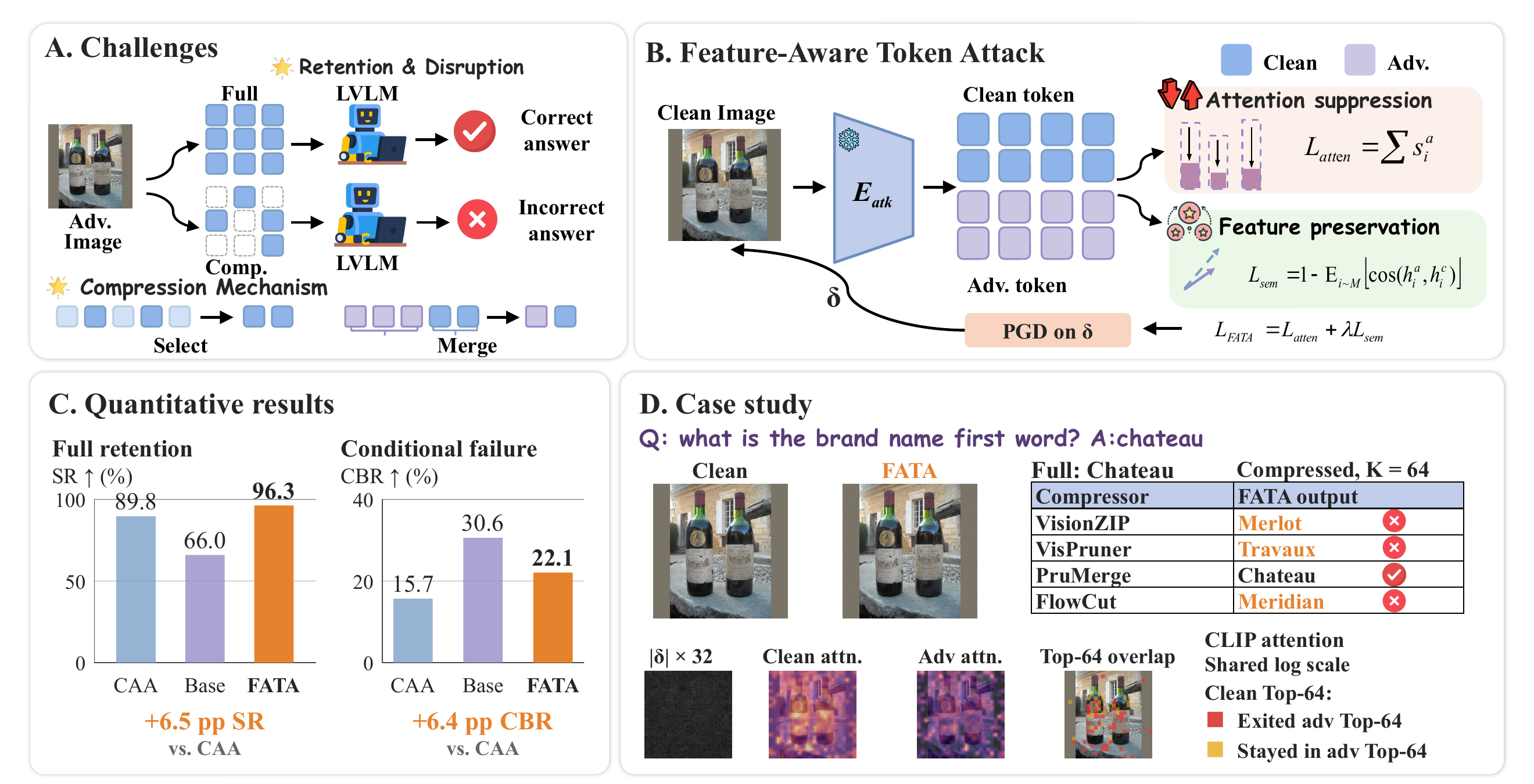}
    \caption{\textbf{FATA: attention suppression with feature preservation.} A illustrates the compression-triggered failure setting; B presents the two attack objectives; C compares full-token retention and conditional blinding; D shows how answers to the same perturbed TextVQA image vary across compression paths.}
    \label{fig:intro-overview}
\end{figure}

We study a failure in which a perturbed image remains correctly understood with the full token sequence but yields an incorrect answer after compression (As shown in Figure~\ref{fig:intro-overview}). When both paths succeed on the clean image, this divergence exposes a vulnerability that full-token evaluation alone would miss. An attack targeting this vulnerability must therefore meet a dual requirement: \emph{induce compressed-path failure while preserving full-token correctness}.

Prior compression-aware attacks approach this problem through feature disruption (CAGE) or token-importance manipulation (CAA) \citep{zhang2026adversarial,zhang2026less}. Related attacks show that vision-encoder access alone can suffice to disrupt visual features \citep{wang2024break,mei2026veattack}. Selective failure poses an additional challenge: encoder attention offers a surrogate for token importance, but perturbing it can also damage the features needed for full-token inference. Conversely, preserving features alone provides little pressure to induce compressed-path failure. This coupling motivates our central question: \textbf{\emph{how can an attack exploit compression sensitivity while preserving full-token behavior, using only vision-encoder access?}}

In this paper, we propose \textbf{Feature-Aware Token Attack (FATA)}, which couples two objectives on a fixed set of tokens selected by clean encoder attention. An \textbf{attention-suppression objective} lowers attention to these positions as a surrogate for compression sensitivity, while a \textbf{feature-preservation objective} maintains cosine alignment with their clean feature vectors. Sharing the target set ties the attack pressure to the content being protected. In the primary LLaVA-1.5-7B setting, FATA optimizes only the input perturbation through a frozen vision encoder, without access to the downstream language model, questions, answers, compressor, or token budget. The same perturbed image is then evaluated across compression rules and budgets without re-optimization.

Across four visually dependent task subsets and four compressors under a controlled reconstruction protocol, FATA achieves \(\SR=\mathbf{96.3\%}\) full-token accuracy retention and \(\CBR=\mathbf{22.1\%}\) conditional blinding rate, compared with \(89.8\%\) and \(15.7\%\) for CAA. Ablations show why both objectives matter: attention suppression alone substantially degrades full-token accuracy, while feature preservation alone yields weaker conditional blinding. These results support jointly controlling attention and feature distortion to expose compression sensitivity while preserving full-token behavior.

The contributions of this paper are summarized as follows.
\begin{itemize}
    \setlength{\itemsep}{2pt}
    \item[\ding{182}] \textbf{Feature-aware attack design.} We introduce FATA, coupling attention suppression with explicit feature preservation on fixed clean-image targets to induce compression-triggered failures under vision-encoder-only access.
    \item[\ding{183}] \textbf{Paired robustness evaluation.} We evaluate the balance between full-token retention and conditional compressed-path failure across four task settings and four compressors, with loss ablations and model-specific extensions under stated access assumptions.
    \item[\ding{184}] \textbf{Detection across model stages.} We introduce a Multi-Level Activation Trajectory Detector (ML-ATD) to monitor the vision encoder, projector, and language model, and compare three detection approaches at a fixed low false-positive rate.
\end{itemize}

%% file: 3-related_work.tex
\section{Related Work}
\label{sec:related_work}
\paragraph{Visual Token Compression}
Inputs at high resolution can produce long visual prefixes, increasing prefill latency and memory use \citep{chen2024image,arif2025hired}. Visual token compression reduces these costs at the encoder output, visual projector, or decoder \citep{zhang2024sparsevlm,li2025tokenpacker,ye2025atp}. Methods that require no training retain dominant, salient, or diverse tokens \citep{yang2025visionzip,zhang2025beyond,alvar2025divprune}. Other approaches merge nearby features or use early decoder attention and information flow across layers to remove redundancy \citep{shang2025llava,chen2024image,tong2026flowcut}.
However, visual token compression can also increase vulnerability to adversarial perturbations that change which tokens are retained for inference \citep{zhang2026adversarial,zhang2026less}.

\paragraph{Adversarial Perturbations under Compression}
Earlier methods disrupt encoder features, alignment between modalities, or transferable representations \citep{wang2024break,lu2023set,zhang2025anyattack}. Recent works use targets generated by diffusion models, encoder objectives independent of downstream tasks, or guidance from projector features \citep{guo2024efficient,mei2026veattack,cao2026enhancing}. These approaches do not model which tokens a compressor retains. For visual token compression, CAGE targets tokens likely to survive unknown budgets \citep{zhang2026adversarial}. Though CAA and our work both focus on inducing failures after compression while preserving full token behavior \citep{zhang2026less}, our method further incorporates an explicit semantic preservation loss to better achieve this goal.


\paragraph{Adversarial Detection}
Adversarial detectors distinguish adversarial inputs from clean inputs. Feature Squeezing assesses prediction consistency under input transformations, while CIDER measures changes in similarity between modalities under denoising \citep{xu2017feature,xu2024cross}. Internal-state detectors use feature distribution deviations or activations associated with safety \citep{lee2018simple,jiang2025hiddendetect}. JailDAM uses adaptive memory to update unsafe knowledge during inference and improve detection of unseen jailbreak strategies \citep{nian2025jaildam}. Alongside input and feature distribution analysis, inspired by HiddenDetect, ML-ATD uses an attack reference direction estimated from Base examples and monitors activations in the visual encoder, projector, and language model.

%% file: 5-method.tex
\section{Method}

\subsection{Overview}

\begin{figure}[H]
    \centering
    \includegraphics[width=\linewidth]{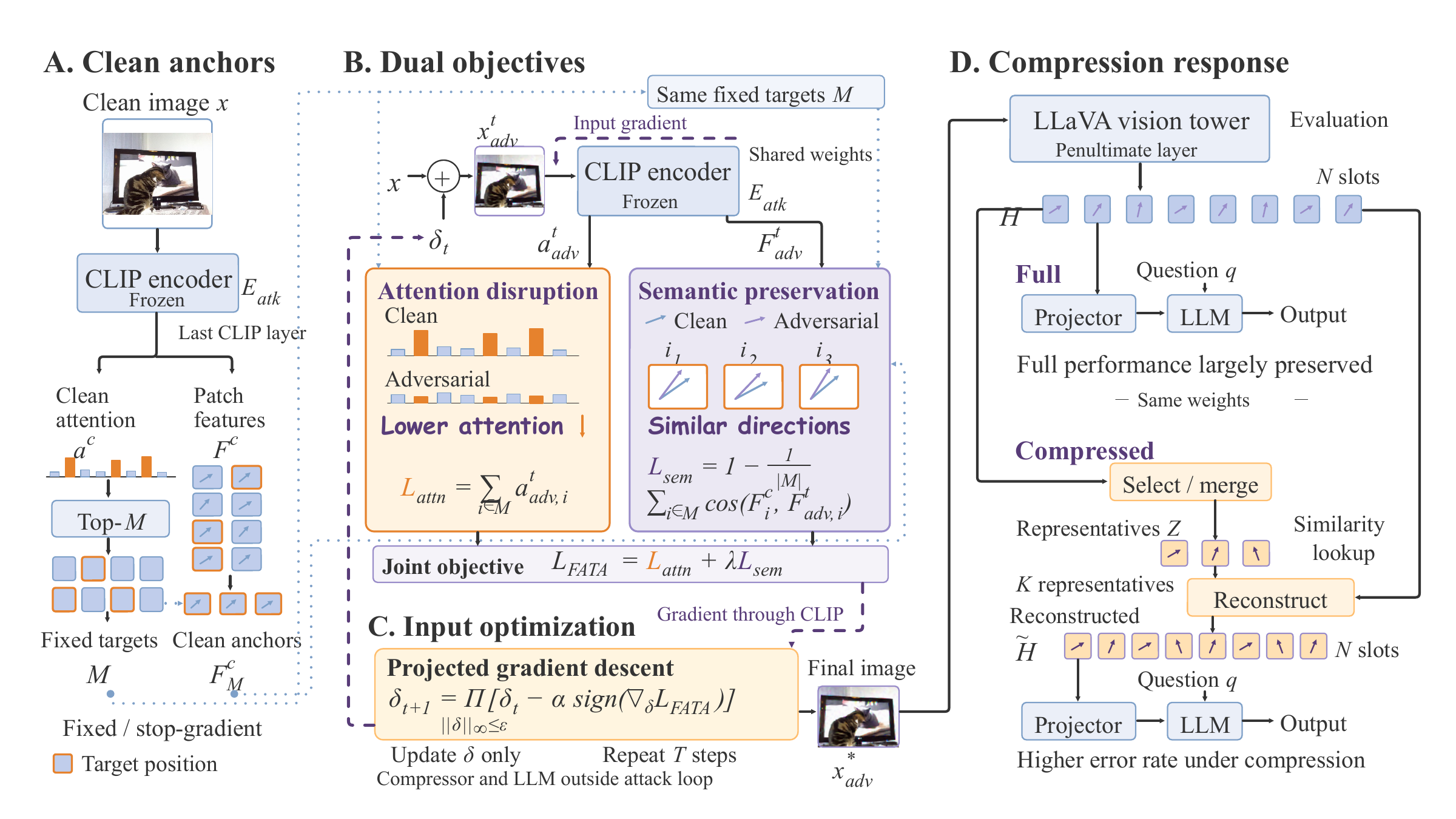}
    \caption{FATA overview. Fixed clean targets guide input-only PGD with attention and semantic losses (A--C), followed by full-token and compressed-path evaluation of the same image (D).}
    \label{fig:fata-overview}
\end{figure}

Figure~\ref{fig:fata-overview} separates attack optimization (A--C) from compression-response evaluation (D). The frozen CLIP encoder uses last-layer clean CLS-to-patch attention to fix the Top-$M$ target tokens and stores their patch features as stop-gradient anchors (A). For each adversarial iterate, $\Lattn$ suppresses attention over those same positions, while $\Lsem$ preserves their feature directions through cosine similarity to the clean anchors (B). PGD backpropagates their weighted sum through CLIP and updates only $\delta$; the encoder weights, target set, and anchors stay fixed, and the compressor and LLM remain outside the attack loop (C).

Panel D evaluates the same optimized image along two paths. LLaVA's vision tower shares the encoder weights but reads penultimate-layer features $H$, distinct from the last-layer attack features. The full-token path passes $H$ directly to the projector. The compressed path selects or merges $K$ representatives $Z$, then uses similarity lookup from $H$ to reconstruct an $N$-slot sequence $\widetilde H$ from those representatives. Both paths use the same projector, LLM, and question. This comparison exposes compression-triggered errors despite largely preserved full-token performance. Reconstruction fills all $N$ positions with representative features.

\Needspace{23\baselineskip}
\subsection{FATA}
\captionsetup{hypcap=false}
\begin{wrapfigure}{r}{0.49\textwidth}
    \vspace{-8pt}
    \begin{minipage}{\linewidth}
\begin{algorithm}[H]
\caption{FATA optimization overview}
\label{alg:fata}
\scriptsize
\begin{algorithmic}[1]
\setlength{\itemsep}{1.5pt}
\Require Clean image $x$, encoder $E$, $\epsilon,\alpha,T,\lambda$, target count $M$
\State $(\mathbf{s}^{c},\mathbf{H}^{c})\gets E(x)$
\State $\mathcal{M}\gets\TopK(\mathbf{s}^{c},M)$
\State $\delta^{(0)}\sim\mathcal U(-\epsilon,\epsilon)$
\For{$t=0,\ldots,T-1$}
  \State $x_a^t\gets\Clip(x+\delta^{(t)},0,1)$
  \State $(\mathbf{s}^{a},\mathbf{H}^{a})\gets E(x_a^t)$
  \State $\Lattn\gets\sum_{i\in\mathcal M}s_i^a$
  \State $\Lsem\gets1-\frac{1}{|\mathcal M|}\sum_{i\in\mathcal M}\cos(\mathbf h_i^a,\mathbf h_i^c)$
  \State $\Ltotal\gets\Lattn+\lambda\Lsem$
  \State $\delta^{(t+1)}\gets\Proj(\delta^{(t)}-\alpha\operatorname{sign}(\nabla_\delta\Ltotal))$
\EndFor
\State \Return $\Clip(x+\delta^{(T)},0,1)$
\end{algorithmic}
\end{algorithm}
    \end{minipage}
    \vspace{10pt}
\end{wrapfigure}
\captionsetup{hypcap=true}

FATA is formulated as a gray-box method on LLaVA with gradient access to the visual encoder $E$, while the compressor, token budget, and downstream language model are unknown during optimization. Given a clean image $x$, FATA optimizes a perturbation $\delta$ under $\|\delta\|_\infty\leq\epsilon$ to obtain $x_{\mathrm{adv}}=\Clip(x+\delta,0,1)$. Let $(\mathbf{s}^{c},\mathbf{H}^{c})=E(x)$ denote the last-layer CLS attention scores over visual tokens and their corresponding features. Attention scores are averaged across heads, and the top $M$ visual tokens form a fixed target set.
\begin{equation}
    \mathcal M=\TopK(\mathbf{s}^{c},M),\qquad M=64.
    \label{eq:target-mask}
\end{equation}
The visual tokens in $\mathcal M$ are referred to as target tokens. For the adversarial image $x_{\mathrm{adv}}$, FATA minimizes the following attention loss over the target tokens.
\begin{equation}
    \Lattn=\sum_{i\in\mathcal M}s_i^{a},
    \label{eq:lattn}
\end{equation}
where $\mathbf{s}^{a}$ denotes the adversarial attention scores. Minimizing $\Lattn$ suppresses attention to the target tokens, with the aim of reducing their selection priority during visual token compression.

To preserve full token behavior, FATA aligns the feature directions of the adversarial target tokens with their clean features. Let $\mathbf h_i^c$ and $\mathbf h_i^a$ denote the clean and adversarial features of target token $i$. The semantic loss is
\begin{equation}
    \Lsem=1-\frac{1}{|\mathcal M|}\sum_{i\in\mathcal M}
    \cos(\mathbf h_i^{a},\mathbf h_i^{c}).
    \label{eq:lsem}
\end{equation}
Both losses use the same fixed target set. The final objective is
\begin{equation}
    \Ltotal=\Lattn+\lambda\Lsem.
    \label{eq:total-loss}
\end{equation}

Algorithm~\ref{alg:fata} applies $\ell_\infty$ PGD to $\Ltotal$ with $\delta^{(0)}\sim\mathcal U(-\epsilon,\epsilon)$, $\lambda=1$, $\epsilon=2/255$, $\alpha=0.5/255$, and $T=100$ steps. Here, $\Proj$ projects the perturbation onto the $\ell_\infty$ ball of radius $\epsilon$. Optimization details and sensitivity to $\alpha$ and $\lambda$ are provided in Appendix~\ref{app:method-details}.

\par\WFclear

\subsection{Evaluation Metrics}
\label{sec:preliminaries}

Full token behavior and compression induced failure are evaluated with SR, CER, and CBR. 
Let $\mathrm{Acc}_{m@K}$ denote the task accuracy of method $m$ at token budget $K$, and let $\mathrm{Acc}_{m@\mathrm{full}}$ denote its accuracy under full token inference. SR measures full token performance retention and is then defined as
\begin{equation}
    \SR=\frac{\mathrm{Acc}_{\mathrm{FATA}@\mathrm{full}}}{\mathrm{Acc}_{\mathrm{Clean}@\mathrm{full}}}.
    \label{eq:sr}
\end{equation}
For LLaVA, the practical token budget $\Kprac$ is selected separately for each dataset--compressor pair using clean accuracy and is fixed across attacks. Qwen3.5 and InternVL3.5 use predefined task-specific retention fractions, as detailed in Appendix~\ref{app:token-budgets}.
\begin{equation}
    \Kprac=\min\left\{K:\frac{\mathrm{Acc}_{\mathrm{Clean}@K}}{\mathrm{Acc}_{\mathrm{Clean}@\mathrm{full}}}\geq0.8\right\}.
    \label{eq:practical-budget}
\end{equation}
For attack $m$, the compressed error rate (CER) measures overall task error at $\Kprac$:
\begin{equation}
    \CER_m=1-\mathrm{Acc}_{m@\Kprac}.
    \label{eq:cer}
\end{equation}
It uses task-credit accuracy and does not subtract the error on Clean inputs.
CBR measures the fraction of examples on which attack $m$ becomes incorrect at the practical budget, conditioned on both Clean and attack $m$ being correct at full tokens and Clean remaining correct under the same compressor and practical budget. Binary correctness follows the task-specific rules in Appendix~\ref{app:metrics}. For FATA, $m=\mathrm{FATA}$,
\begin{equation}
    \CBR=
    \frac{\left|\left\{x:\begin{aligned}
    &\mathrm{Clean}@\mathrm{full}\ \text{correct}\ \land\ m@\mathrm{full}\ \text{correct}\ \land\\
    &\mathrm{Clean}@\Kprac\ \text{correct}\ \land\ m@\Kprac\ \text{incorrect}
    \end{aligned}\right\}\right|}
    {\left|\left\{x:\begin{aligned}
    &\mathrm{Clean}@\mathrm{full}\ \text{correct}\ \land\ m@\mathrm{full}\ \text{correct}\ \land\\
    &\mathrm{Clean}@\Kprac\ \text{correct}
    \end{aligned}\right\}\right|}.
    \label{eq:cbr}
\end{equation}

Scoring rules and metric aggregation across the three models are detailed in Appendix~\ref{app:metrics}.

%% file: 6-experiments.tex
\section{Experiments}

\subsection{Experimental Setup}

The main evaluation uses LLaVA-1.5-7B \citep{liu2024improved} with four visual token compressors, VisionZIP, VisPruner, PruMerge, and FlowCut \citep{yang2025visionzip,zhang2025beyond,shang2025llava,tong2026flowcut}. The evaluation covers four 1,000-image subsets, TextVQA-Open, VQAv2-Open, ScienceQA-MC, and VQAv2-MC \citep{singh2019textvqa,goyal2017vqav2,lu2022scienceqa}. Before inference, one representative question is selected for each unique image. The subset is then constructed by retaining examples that are answered correctly with the original image but incorrectly when the image is replaced by a black image. Further details of subset construction are provided in Appendix~\ref{app:implementation}. The retained token budgets are $K\in\{576,192,128,64,32,16\}$, and FATA is compared with Base(attention-only), CAA \citep{zhang2026less}, and CAGE \citep{zhang2026adversarial} under the common perturbation bound $\epsilon=2/255$.

For each dataset and compressor, clean and adversarial inputs are evaluated on the same valid examples at every retained token budget using the metrics defined in Section~\ref{sec:preliminaries}. $K_{\mathrm{prac}}$ is selected from clean results only and then fixed for attack evaluation, as defined in Eq.~\ref{eq:practical-budget}.

\subsection{LLaVA-1.5-7B Results}
\subsubsection{Main Results}

Table~\ref{tab:main-results} reports results on LLaVA-1.5-7B \citep{liu2024improved} and shows that FATA preserves most full token performance while causing larger degradation under visual token compression.

At $K=576$, FATA achieves an average accuracy of 89.7\%, close to the 93.2\% Clean average, while Base and CAGE fall to 61.6\% and 34.4\%, respectively. This result indicates that FATA preserves full token behavior before compression is applied.

At matched compression budgets, the accuracy difference between Clean and FATA increases from 3.5 points at $K=576$ to 5.5, 7.4, 9.3, 12.5, and 15.6 points as the retained token budget decreases to 192, 128, 64, 32, and 16, respectively. At $K=16$, TextVQA-Open exhibits the largest dataset level difference, at 23.8 points. The widening gap shows that FATA induces larger accuracy degradation after visual token compression.

\input{tables/table1_main_results}

\begin{wrapfigure}[11]{r}{0.49\textwidth}
    \vspace{-8pt}
    \centering
    \includegraphics[width=\linewidth]{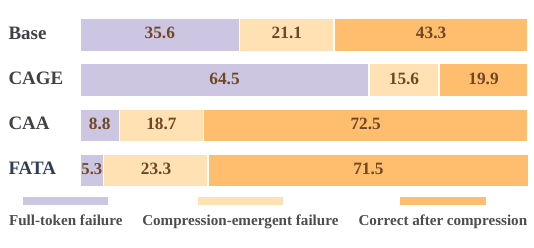}
    \captionsetup{font=small,skip=2pt}
    \caption{Failure-stage decomposition over 14,316 shared Clean-full-correct cases.}
    \label{fig:error-stage-decomposition}
    \vspace{-8pt}
\end{wrapfigure}

Figure~\ref{fig:error-stage-decomposition} shows when failures occur across attacks using the same examples answered correctly on clean inputs at full tokens. Among the evaluated methods, FATA has the fewest failures before compression (5.3\%) and the most emerging after compression (23.3\%), while 71.5\% of cases remain correct after compression. CAA also limits failures before compression (8.8\%) but has fewer failures emerging after compression (18.7\%). Compared with FATA, Base and CAGE have more failures before compression, at 35.6\% and 64.5\%, respectively. These percentages share the Clean-full-correct denominator. CBR conditions on Clean full-token, attacked full-token, and Clean compressed correctness, then measures the attack error rate under the same compressor and practical budget.

\WFclear

\Needspace{13\baselineskip}
\input{tables/table2_stealth_lethality}

Table~\ref{tab:stealth-lethality} reports SR, CER, and CBR to characterize the balance between full token retention and failure after compression at $K_{\mathrm{prac}}$. CER is computed as $100\%-\mathrm{Acc}@K_{\mathrm{prac}}$ from the displayed accuracies in Table~\ref{tab:main-results} and budgets in Table~\ref{tab:llava-kprac}, then averaged over the 16 settings and rounded to two decimal places.

Compared with Base, FATA raises SR from 66.7\% to 96.3\% while reducing CER from 54.78\% to 30.38\% and CBR from 32.9\% to 22.1\%, balancing full token behavior and compression induced failure. CAGE reaches higher CER and CBR, 75.58\% and 43.1\%, respectively, but its SR drops to 36.6\%, while CAA keeps 89.8\% SR but produces a lower CBR of 15.7\%. These results indicate that Base and CAGE exhibit higher CER and CBR alongside greater full-token degradation, while FATA preserves full-token performance more effectively. Compared with CAA, FATA increases SR from 89.8\% to 96.3\% and CBR from 15.7\% to 22.1\%, offering a better trade-off. Appendix~\ref{app:qualitative-cases} provides qualitative examples of changes in CLIP attention rankings and compression-dependent answers.

\par\WFclear

\subsubsection{Ablation Study}
\input{tables/table3_dual_objective_ablation}
Table~\ref{tab:ablation} separates the roles of the two losses in Eq.~\ref{eq:total-loss}. The attention loss $\Lattn$ alters the visual evidence that survives compression, whereas the semantic loss $\Lsem$ preserves the full token representation. Applying $\Lattn$ alone gives the highest CBR of 32.9\% but reduces SR to 66.7\%, showing that attention disruption strengthens compression induced failure but damages full token behavior without semantic preservation. Using $\Lsem$ alone gives the opposite effect, reaching 98.5\% SR but only 14.5\% CBR, which indicates that semantic preservation protects full token inference but leads to fewer compression-induced failures. The full FATA objective maintains 96.3\% SR and reaches 22.1\% CBR, combining the full token behavior supported by $\Lsem$ with the compression sensitivity introduced by $\Lattn$. Appendix~\ref{app:ablation} shows the corresponding variation across the 16 experimental settings.

\subsection{Cross-Model Evaluation}
\label{sec:cross-model}
\subsubsection{Qwen3.5-9B}

Since the number of visual tokens in Qwen3.5\citep{qwen2026qwen35} depends on input resolution, compression budgets are specified as fractions of visual tokens. The evaluation uses a $1/3$ token budget, while the practical token budget is set to $1/9$ for open tasks and $1/18$ for multiple choice tasks. Model-specific adaptations are detailed in Appendix~\ref{app:qwen-adaptation}.

\input{tables/table4_qwen_accuracy}

\Needspace{18\baselineskip}
\input{fig_tex/qwen_cbr}

Figure~\ref{fig:qwen-cbr} shows the corresponding CBR values for each dataset and compressor.
At the practical token budget, FATA accuracy decreases by 40.44 percentage points, from 88.92\% to 48.48\%, while Clean accuracy decreases by 24.45 points, from 89.55\% to 65.10\%, as reported in Table~\ref{tab:qwen}. The larger decrease for FATA shows that compression leads to greater accuracy loss with FATA on Qwen3.5.

Table~\ref{tab:qwen-practical-cbr} reports CBR and SR on Qwen3.5-9B. SR remains high at an average of 99.3\% across the four datasets, while CBR averages 46.43\% across the 16 dataset and compressor combinations. These results support the same pattern observed on LLaVA, with FATA retaining full token behavior while exhibiting failures after compression.

\par\WFclear

\input{tables/table5_qwen_sr_cbr}

\subsubsection{InternVL3.5-8B-HF}

To accommodate the dynamic tiling and token organization of InternVL3.5-8B-HF \citep{wang2025internvl35advancingopensourcemultimodal}, \emph{TA-FATA} adapts the FATA loss while retaining attention disruption and semantic preservation. The complete objective, fixed hyperparameters, and evaluation details are given in Appendix~\ref{app:internvl-adaptation}.

Table~\ref{tab:internvl} reports the accuracy of TA-FATA across token budgets. At full tokens, the four-task average accuracy is 90.52\% for Clean and 85.28\% for TA-FATA, while the corresponding values at the $1/3$ budget are 85.96\% and 75.59\%. At the practical budget, TA-FATA accuracy decreases by 32.81 percentage points from full tokens to 52.47\%, while Clean accuracy decreases by 24.08 points to 66.44\%. As on LLaVA and Qwen3.5, visual token compression leads to greater accuracy loss with TA-FATA than with Clean inputs.

\input{tables/table6_internvl_accuracy}

\input{tables/table7_internvl_sr_cbr}

Table~\ref{tab:internvl-practical-cbr} reports exact-match SR and conditional CBR on InternVL3.5-8B-HF. Macro SR is 93.94\%, compared with 96.3\% on LLaVA and 99.3\% on Qwen3.5. Macro CBR over the 16 dataset and compressor combinations is 40.38\%, higher than LLaVA's 22.1\% but lower than Qwen3.5's 46.43\%. InternVL3.5 therefore incurs a greater cost to full-token preservation while exhibiting a conditional blinding rate between the other two models. Its exact-match SR uses binary correctness, while Table~\ref{tab:internvl} reports canonical task-credit accuracy, so the latter cannot directly reproduce SR.

Across LLaVA, Qwen3.5, and InternVL3.5, FATA and its adaptations retain most full token performance while exhibiting greater accuracy loss under compression than Clean.

%% file: tables/table1_main_results.tex
\begin{table*}[t!]
\centering
\caption{Accuracy (\%) across four benchmarks and visual-token budgets.}
\label{tab:main-results}
\begingroup
\scriptsize
\setlength{\tabcolsep}{1.15pt}
\renewcommand{\arraystretch}{0.94}
\setlength{\aboverulesep}{0pt}
\setlength{\belowrulesep}{0pt}
\setlength{\cmidrulesep}{0pt}
\resizebox{\textwidth}{!}{%
\begin{tabular}{l|ccccc|ccccc|ccccc|ccccc}
\toprule
\rowcolor{HeaderGray}&
\multicolumn{5}{c|}{\textbf{TextVQA-Open}} &
\multicolumn{5}{c|}{\textbf{VQAv2-Open}} &
\multicolumn{5}{c|}{\textbf{ScienceQA-MC}} &
\multicolumn{5}{c}{\textbf{VQAv2-MC}}\\
\rowcolor{HeaderGray}\multirow{-2}{*}{\textbf{Method}} & Clean & Base & CAA & CAGE & \textbf{FATA} & Clean & Base & CAA & CAGE & \textbf{FATA} & Clean & Base & CAA & CAGE & \textbf{FATA} & Clean & Base & CAA & CAGE & \textbf{FATA}\\
\midrule
\multicolumn{21}{c}{\cellcolor{gray!20}\rule[-0.70ex]{0pt}{2.55ex}\textit{Upper Bound (576 Tokens)}}\\
None & 92.0 & 39.9 & 77.4 & 19.8 & 86.6 & 88.2 & 66.1 & 78.9 & 28.3 & 86.0 & 94.5 & 57.8 & 89.2 & 40.6 & 90.1 & 98.0 & 82.5 & 89.6 & 48.7 & 96.2\\
\midrule
\multicolumn{21}{c}{\cellcolor{gray!20}\rule[-0.70ex]{0pt}{2.55ex}\textit{Retain 192 Tokens ($K_{model}=192$)}}\\
VisionZIP & 89.1 & 35.7 & 82.5 & 16.2 & 83.3 & 86.6 & 61.8 & 56.2 & 25.8 & 84.2 & 91.3 & 54.5 & 86.3 & 38.2 & 86.3 & 96.9 & 79.1 & 94.8 & 45.8 & 95.1\\
VisPruner & 90.6 & 36.1 & 83.9 & 17.9 & 83.3 & 87.4 & 65.1 & 84.9 & 26.6 & 84.3 & 91.1 & 56.5 & 87.1 & 40.2 & 88.7 & 96.8 & 79.8 & 95.1 & 45.0 & 95.0\\
FlowCut & 88.1 & 38.0 & 81.9 & 18.5 & 82.9 & 86.2 & 63.2 & 84.6 & 24.1 & 83.3 & 87.7 & 56.5 & 86.1 & 39.3 & 87.0 & 96.0 & 77.1 & 94.6 & 45.5 & 94.1\\
PruMerge & 84.5 & 29.8 & 77.7 & 16.3 & 60.4 & 84.1 & 58.5 & 83.3 & 25.5 & 79.1 & 89.4 & 50.3 & 86.9 & 38.1 & 74.9 & 94.8 & 75.3 & 65.3 & 42.8 & 90.7\\
\rowcolor{avgrow}\rule[-0.30ex]{0pt}{1.85ex}\textbf{Average} & 88.1 & 34.9 & 81.5 & 17.2 & \textbf{77.5} & 86.1 & 62.2 & 77.3 & 25.5 & \textbf{82.7} & 89.9 & 54.5 & 86.6 & 39.0 & \textbf{84.2} & 96.1 & 77.8 & 87.5 & 44.8 & \textbf{93.7}\\
\midrule
\multicolumn{21}{c}{\cellcolor{gray!20}\rule[-0.70ex]{0pt}{2.55ex}\textit{Retain 128 Tokens ($K_{model}=128$)}}\\
VisionZIP & 86.7 & 32.3 & 81.6 & 13.5 & 80.2 & 86.1 & 58.5 & 58.4 & 22.6 & 82.7 & 88.9 & 53.5 & 85.5 & 37.1 & 85.3 & 95.5 & 76.3 & 93.3 & 43.0 & 92.9\\
VisPruner & 88.2 & 34.8 & 81.8 & 15.9 & 81.3 & 86.0 & 62.3 & 84.2 & 23.4 & 82.9 & 89.4 & 56.3 & 85.7 & 39.3 & 87.1 & 95.5 & 76.3 & 94.7 & 42.9 & 93.7\\
FlowCut & 84.9 & 32.8 & 79.4 & 15.5 & 78.4 & 84.0 & 58.5 & 82.3 & 19.6 & 80.7 & 85.6 & 56.6 & 83.6 & 39.5 & 84.6 & 94.0 & 72.6 & 92.7 & 39.7 & 90.5\\
PruMerge & 81.8 & 23.9 & 76.6 & 12.9 & 46.0 & 80.6 & 54.4 & 79.9 & 23.5 & 69.8 & 88.5 & 47.6 & 85.9 & 36.2 & 71.5 & 92.6 & 71.2 & 67.6 & 41.1 & 83.0\\
\rowcolor{avgrow}\rule[-0.30ex]{0pt}{1.85ex}\textbf{Average} & 85.4 & 30.9 & 79.9 & 14.5 & \textbf{71.5} & 84.2 & 58.4 & 76.2 & 22.3 & \textbf{79.0} & 88.1 & 53.5 & 85.2 & 38.0 & \textbf{82.1} & 94.4 & 74.1 & 87.1 & 41.7 & \textbf{90.0}\\
\midrule
\multicolumn{21}{c}{\cellcolor{gray!20}\rule[-0.70ex]{0pt}{2.55ex}\textit{Retain 64 Tokens ($K_{model}=64$)}}\\
VisionZIP & 81.2 & 25.8 & 75.7 & 11.4 & 70.9 & 80.3 & 51.4 & 58.8 & 20.1 & 77.3 & 85.3 & 52.0 & 82.8 & 35.6 & 83.1 & 90.7 & 68.4 & 87.7 & 37.0 & 86.3\\
VisPruner & 82.6 & 29.7 & 75.6 & 13.1 & 74.6 & 80.3 & 55.8 & 78.6 & 20.9 & 75.3 & 85.9 & 53.4 & 83.8 & 39.7 & 84.1 & 92.2 & 71.4 & 89.8 & 35.6 & 88.7\\
FlowCut & 76.8 & 24.6 & 72.4 & 11.2 & 66.8 & 74.1 & 47.7 & 72.6 & 14.5 & 69.7 & 82.4 & 51.8 & 80.6 & 37.1 & 76.7 & 85.5 & 60.8 & 85.9 & 31.0 & 79.8\\
PruMerge & 77.7 & 20.0 & 73.6 & 10.4 & 38.2 & 77.8 & 47.7 & 75.8 & 20.5 & 62.5 & 84.8 & 46.4 & 85.5 & 34.7 & 66.4 & 90.6 & 65.7 & 66.4 & 38.0 & 78.5\\
\rowcolor{avgrow}\rule[-0.30ex]{0pt}{1.85ex}\textbf{Average} & 79.6 & 25.0 & 74.3 & 11.5 & \textbf{62.6} & 78.1 & 50.6 & 71.4 & 19.0 & \textbf{71.2} & 84.6 & 50.9 & 83.2 & 36.8 & \textbf{77.6} & 89.8 & 66.6 & 82.4 & 35.4 & \textbf{83.3}\\
\midrule
\multicolumn{21}{c}{\cellcolor{gray!20}\rule[-0.70ex]{0pt}{2.55ex}\textit{Retain 32 Tokens ($K_{model}=32$)}}\\
VisionZIP & 69.0 & 17.0 & 64.8 & 8.8 & 54.7 & 67.6 & 37.4 & 51.5 & 15.1 & 61.2 & 81.6 & 47.2 & 79.9 & 32.2 & 75.2 & 82.3 & 54.9 & 79.8 & 32.3 & 74.9\\
VisPruner & 72.7 & 22.9 & 68.2 & 10.0 & 60.1 & 70.5 & 44.0 & 68.3 & 16.7 & 64.6 & 79.3 & 51.0 & 75.6 & 39.8 & 74.1 & 84.7 & 58.4 & 83.2 & 32.4 & 79.0\\
FlowCut & 63.0 & 16.7 & 59.3 & 7.9 & 46.9 & 58.7 & 32.9 & 57.6 & 10.5 & 49.8 & 79.8 & 47.6 & 78.8 & 35.2 & 72.6 & 73.9 & 48.5 & 71.5 & 26.5 & 63.2\\
PruMerge & 66.3 & 14.4 & 61.4 & 6.6 & 26.7 & 66.1 & 37.6 & 63.5 & 16.0 & 51.1 & 79.7 & 44.5 & 76.6 & 33.7 & 55.9 & 81.7 & 56.5 & 62.7 & 32.0 & 67.1\\
\rowcolor{avgrow}\rule[-0.30ex]{0pt}{1.85ex}\textbf{Average} & 67.8 & 17.8 & 63.4 & 8.3 & \textbf{47.1} & 65.7 & 38.0 & 60.2 & 14.6 & \textbf{56.7} & 80.1 & 47.6 & 77.7 & 35.2 & \textbf{69.5} & 80.7 & 54.6 & 74.3 & 30.8 & \textbf{71.1}\\
\midrule
\multicolumn{21}{c}{\cellcolor{gray!20}\rule[-0.70ex]{0pt}{2.55ex}\textit{Retain 16 Tokens ($K_{model}=16$)}}\\
VisionZIP & 53.3 & 12.1 & 49.2 & 6.1 & 30.6 & 52.5 & 26.4 & 42.8 & 12.4 & 38.6 & 69.9 & 39.3 & 64.9 & 32.4 & 54.2 & 70.2 & 41.6 & 67.2 & 26.4 & 55.4\\
VisPruner & 54.2 & 14.4 & 49.2 & 8.2 & 31.0 & 52.4 & 29.9 & 52.8 & 12.4 & 41.8 & 66.7 & 46.2 & 67.2 & 34.9 & 61.5 & 70.9 & 51.0 & 67.8 & 29.2 & 61.7\\
FlowCut & 44.2 & 10.8 & 41.1 & 5.5 & 23.8 & 39.7 & 18.7 & 38.8 & 10.4 & 25.5 & 64.0 & 40.4 & 56.4 & 33.9 & 47.9 & 55.8 & 33.5 & 54.5 & 26.1 & 40.3\\
PruMerge & 45.9 & 10.1 & 44.5 & 6.2 & 17.1 & 49.3 & 27.6 & 51.5 & 13.3 & 37.0 & 62.5 & 42.3 & 58.0 & 33.4 & 48.6 & 69.3 & 46.4 & 52.6 & 28.7 & 55.6\\
\rowcolor{avgrow}\rule[-0.30ex]{0pt}{1.85ex}\textbf{Average} & 49.4 & 11.9 & 46.0 & 6.5 & \textbf{25.6} & 48.5 & 25.7 & 46.5 & 12.1 & \textbf{35.7} & 65.8 & 42.1 & 61.6 & 33.7 & \textbf{53.1} & 66.6 & 43.1 & 60.5 & 27.6 & \textbf{53.2}\\
\bottomrule
\end{tabular}}
\endgroup
\end{table*}

%% file: tables/table2_stealth_lethality.tex
\begin{wraptable}{r}{0.46\textwidth}
\vspace{-6pt}
\centering
\captionsetup{hypcap=false}
\captionof{table}{Overall stealth--lethality trade-off}
\label{tab:stealth-lethality}
\vspace{2pt}
\scriptsize
\renewcommand{\arraystretch}{1.10}
\setlength{\tabcolsep}{4.0pt}
\resizebox{\linewidth}{!}{%
\begin{tabular}{lccc}
\toprule
\rowcolor{HeaderGray}
\textbf{Attack} & \textbf{SR $\uparrow$} & \textbf{CER $\uparrow$} & \textbf{CBR $\uparrow$}\\
\midrule
Base (Attn.-only) & 66.7 & 54.78 & 32.9\\
CAA & 89.8 & 24.65 & 15.7\\
CAGE & 36.6 & \textbf{75.58} & \textbf{43.1}\\
\rowcolor{fatarow}\textbf{FATA} & \textbf{96.3} & 30.38 & 22.1\\
\bottomrule
\end{tabular}
}
\vspace{-6pt}
\end{wraptable}

%% file: tables/table3_dual_objective_ablation.tex
\begin{wraptable}{r}{0.46\textwidth}
\vspace{-6pt}
\centering
\captionsetup{hypcap=false}
\captionof{table}{Dual-objective ablation of FATA.}
\label{tab:ablation}
\vspace{2pt}
\scriptsize
\renewcommand{\arraystretch}{1.10}
\setlength{\tabcolsep}{4.0pt}
\resizebox{\linewidth}{!}{%
\begin{tabular}{lccc}
\toprule
\rowcolor{HeaderGray}
\textbf{Variant} & \textbf{Objective} & \textbf{SR $\uparrow$} & \textbf{CBR $\uparrow$}\\
\midrule
Attention-only & $\Lattn$ & 66.7 & \textbf{32.9}\\
Semantic-only & $\Lsem$ & \textbf{98.5} & 14.5\\
\rowcolor{fatarow}\textbf{FATA Full} & $\Lattn+\lambda\Lsem$ & 96.3 & 22.1\\
\bottomrule
\end{tabular}
}
\vspace{-6pt}
\end{wraptable}

%% file: tables/table4_qwen_accuracy.tex
\Needspace{16\baselineskip}
\begin{center}
    \captionsetup{hypcap=false}
    \captionof{table}{Qwen3.5-9B macro accuracy (\%) under full, $1/3$, and practical token budgets. Values are averaged over four datasets and the four compressors.}
    \label{tab:qwen}
    \small
    \renewcommand{\arraystretch}{1.12}
    \setlength{\tabcolsep}{7pt}
    \begin{tabular}{lrrrr}
        \toprule
        \rowcolor{HeaderGray}
        \textbf{Visual-token budget} & \textbf{Clean} & \textbf{FATA} & \textbf{Damage (pp)} & \textbf{Amplification (pp)} \\
        \midrule
        Full & 89.55 & 88.92 & 0.62 & -- \\
        Retain $1/3$ & 80.66 & 72.47 & 8.19 & 7.56 \\
        \rowcolor{fatarow}Practical (Open: $1/9$; MC: $1/18$) & 65.10 & 48.48 & \textbf{16.62} & \textbf{15.99} \\
        \bottomrule
    \end{tabular}
\end{center}

%% file: fig_tex/qwen_cbr.tex
\begin{wrapfigure}{r}{0.46\textwidth}
    \vspace{-6pt}
    \centering
    \includegraphics[width=\linewidth]{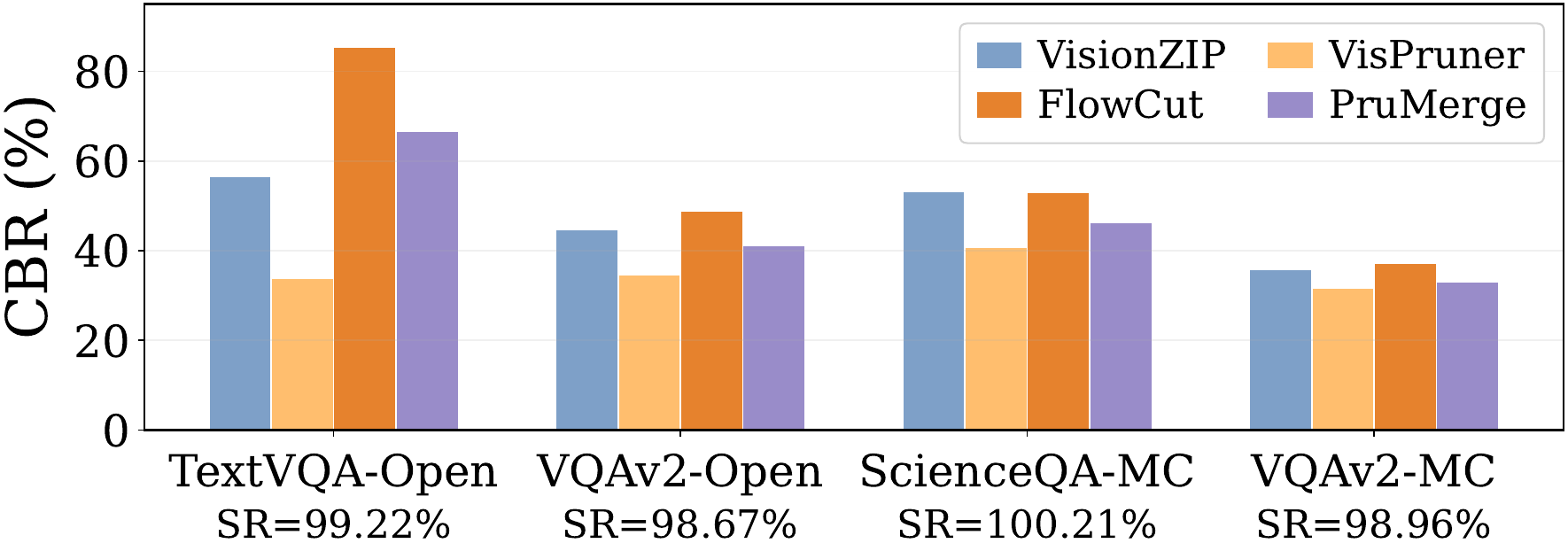}
    \caption{Qwen3.5 practical-budget CBR by compressor and dataset.}
    \label{fig:qwen-cbr}
    \vspace{-6pt}
\end{wrapfigure}

%% file: tables/table5_qwen_sr_cbr.tex
\Needspace{13\baselineskip}
\begin{center}
    \captionsetup{hypcap=false}
    \captionof{table}{Qwen3.5-9B full-token SR and practical-budget conditional CBR (\%) across the four compressors.}
    \label{tab:qwen-practical-cbr}
    \small
    \renewcommand{\arraystretch}{1.10}
    \setlength{\tabcolsep}{4.5pt}
    \begin{tabularx}{\textwidth}{C{2.85cm} Y Y Y Y}
        \toprule
        \rowcolor{HeaderGray}
        \makecell[c]{\textbf{Compression}\\[-0.1ex]\textbf{Method}}
        & \makecell[c]{\textbf{TextVQA-Open}\\\footnotesize SR=99.22\%}
        & \makecell[c]{\textbf{VQAv2-Open}\\\footnotesize SR=98.67\%}
        & \makecell[c]{\textbf{ScienceQA-MC}\\\footnotesize SR=100.21\%}
        & \makecell[c]{\textbf{VQAv2-MC}\\\footnotesize SR=98.96\%} \\
        \midrule
        \rowcolor{BudgetGray}\multicolumn{5}{c}{\textit{CBR at Practical Budget (Open: $1/9$; MC: $1/18$)}} \\
        VisionZIP & 56.58 & 44.70 & 53.28 & 35.85 \\
        VisPruner & 33.79 & 34.66 & 40.79 & 31.66 \\
        FlowCut & 85.37 & 48.83 & 53.07 & 37.21 \\
        PruMerge & 66.67 & 41.13 & 46.20 & 33.02 \\
        \rowcolor{fatarow}\textbf{Average} & \textbf{60.60} & \textbf{42.33} & \textbf{48.34} & \textbf{34.43} \\
        \bottomrule
    \end{tabularx}
\end{center}

%% file: tables/table6_internvl_accuracy.tex
\Needspace{25\baselineskip}
\begin{center}
    \captionsetup{hypcap=false}
    \captionof{table}{InternVL3.5-8B-HF accuracy (\%) under full, $1/3$, and practical token budgets. Compressed rows average the four retained compressors; the practical budget is $1/9$ for open-ended tasks and $1/18$ for multiple-choice tasks.}
    \label{tab:internvl}
    \scriptsize
    \renewcommand{\arraystretch}{1.08}
    \setlength{\tabcolsep}{2.4pt}
    \begin{tabularx}{\textwidth}{C{2.65cm} Y Y Y Y Y Y Y Y}
        \toprule
        \rowcolor{HeaderGray}
        & \multicolumn{8}{c}{\textbf{InternVL3.5-8B-HF Accuracy (\%)}} \\
        \rowcolor{HeaderGray}
        & \multicolumn{2}{c}{\textbf{TextVQA-Open}}
        & \multicolumn{2}{c}{\textbf{VQAv2-Open}}
        & \multicolumn{2}{c}{\textbf{ScienceQA-MC}}
        & \multicolumn{2}{c}{\textbf{VQAv2-MC}} \\
        \rowcolor{HeaderGray}
        \multirow{-3}{*}{\makecell[c]{\textbf{Compression}\\[-0.1ex]\textbf{Method}}}
        & \textbf{Clean} & \textbf{FATA} & \textbf{Clean} & \textbf{FATA}
        & \textbf{Clean} & \textbf{FATA} & \textbf{Clean} & \textbf{FATA} \\
        \midrule
        \rowcolor{BudgetGray}\multicolumn{9}{c}{\textit{Full Visual Tokens}} \\
        No Compression & 87.6 & 81.2 & 79.1 & 70.0 & 98.7 & 96.6 & 96.7 & 93.3 \\
        \midrule
        \rowcolor{BudgetGray}\multicolumn{9}{c}{\textit{Retain $1/3$ of Visual Tokens}} \\
        VisionZIP & 81.6 & 64.6 & 77.8 & \textbf{66.7} & 97.1 & \textbf{91.9} & 96.0 & \textbf{91.2} \\
        VisPruner & 66.0 & 44.0 & 75.9 & 62.2 & 95.2 & 88.6 & 94.5 & 89.6 \\
        FlowCut & 78.4 & 62.0 & 75.8 & 61.9 & 95.0 & 88.8 & 94.5 & 88.5 \\
        PruMerge & 82.3 & \textbf{65.3} & 76.9 & 66.1 & 93.5 & 88.1 & 94.8 & 90.0 \\
        \rowcolor{fatarow}\textbf{Average} & \textbf{77.1} & \textbf{59.0} & \textbf{76.6} & \textbf{64.2} & \textbf{95.2} & \textbf{89.4} & \textbf{95.0} & \textbf{89.8} \\
        \midrule
        \rowcolor{BudgetGray}\multicolumn{9}{c}{\textit{Practical Budget (Open: $1/9$; MC: $1/18$)}} \\
        VisionZIP & 51.8 & 31.9 & 71.1 & 53.1 & 68.1 & 57.4 & 84.5 & 72.8 \\
        VisPruner & 41.6 & 22.8 & 66.2 & 48.2 & 67.9 & 57.5 & 84.6 & 72.7 \\
        FlowCut & 49.2 & 26.8 & 66.5 & 47.6 & 64.5 & 54.8 & 80.0 & 69.4 \\
        PruMerge & 47.3 & \textbf{33.4} & 69.9 & \textbf{56.9} & 67.9 & \textbf{60.7} & 81.9 & \textbf{73.5} \\
        \rowcolor{fatarow}\textbf{Average} & \textbf{47.5} & \textbf{28.7} & \textbf{68.4} & \textbf{51.5} & \textbf{67.1} & \textbf{57.6} & \textbf{82.8} & \textbf{72.1} \\
        \bottomrule
    \end{tabularx}
\end{center}

%% file: tables/table7_internvl_sr_cbr.tex
\Needspace{16\baselineskip}
\begin{center}
    \captionsetup{hypcap=false}
    \captionof{table}{InternVL3.5-8B-HF full-token SR and practical-budget conditional CBR (\%) across four compressors.}
    \label{tab:internvl-practical-cbr}
    \small
    \renewcommand{\arraystretch}{1.10}
    \setlength{\tabcolsep}{4.5pt}
    \begin{tabularx}{\textwidth}{C{2.85cm} Y Y Y Y}
        \toprule
        \rowcolor{HeaderGray}
        \makecell[c]{\textbf{Compression}\\[-0.1ex]\textbf{Method}}
        & \makecell[c]{\textbf{TextVQA-Open}\\\footnotesize SR=93.29\%}
        & \makecell[c]{\textbf{VQAv2-Open}\\\footnotesize SR=88.11\%}
        & \makecell[c]{\textbf{ScienceQA-MC}\\\footnotesize SR=97.87\%}
        & \makecell[c]{\textbf{VQAv2-MC}\\\footnotesize SR=96.48\%} \\
        \midrule
        \rowcolor{BudgetGray}\multicolumn{5}{c}{\textit{CBR at Practical Budget (Open: $1/9$; MC: $1/18$)}} \\
        VisionZIP & 62.18 & 28.00 & 41.39 & 23.00 \\
        VisPruner & 72.48 & 36.07 & 41.18 & 23.54 \\
        FlowCut & 67.64 & 36.61 & 43.88 & 26.57 \\
        PruMerge & 60.12 & 23.96 & 37.55 & 21.92 \\
        \rowcolor{fatarow}\textbf{Average} & \textbf{65.61} & \textbf{31.16} & \textbf{41.00} & \textbf{23.76} \\
        \bottomrule
    \end{tabularx}
\end{center}

%% file: 7-defense.tex
\section{Detection Analysis}

Beyond the performance of FATA in full token retention and failure after compression, this section evaluates the detectability of FATA inputs using input and internal-state detectors.

Table~\ref{tab:detection-cross} reports the area under the receiver operating characteristic curve (AUROC), the area under the precision recall curve (AUPR), and true positive rate at a $5\%$ false positive rate (TPR@5FPR) for FATA, CAA, CAGE, and Base using Feature Squeezing \citep{xu2017feature}, Mahalanobis-Max \citep{lee2018simple}, and a Multi-Level Activation Trajectory Detector (ML-ATD) inspired by HiddenDetect \citep{jiang2025hiddendetect}. Among the three detectors, Feature Squeezing compares model predictions before and after
input transformations by measuring changes in their scores against ground-truth answers. Mahalanobis-Max measures feature space deviation from normal samples; ML-ATD tracks changes in activations across the visual encoder, projector, and LLM. Additional detector construction and evaluation details are provided in Appendix~\ref{app:detection-details}.

\input{tables/table11_detection_cross_attack}

FATA has the lowest TPR@5FPR among the four attacks for each detector. Feature Squeezing gives similarly low recall for FATA and CAA (2.4\% and 2.7\%), whereas Mahalanobis-Max separates them more clearly (13.4\% and 46.8\%). Under ML-ATD, FATA reaches 49.7\%, compared with 53.3\% for CAA, 69.4\% for CAGE, and 80.1\% for Base; CAA has slightly lower AUROC and AUPR. Table~\ref{tab:detection-main} in Appendix~\ref{app:detection-details} reports the relative reductions from Base to FATA.

%% file: tables/table11_detection_cross_attack.tex
\begin{table}[H]
\centering
\captionsetup{skip=3pt}
\caption{Cross-attack detection performance with clean and random negatives. Higher values indicate easier detection; protocol differences and limitations are detailed in Appendix~\ref{app:detection-details}.}
\label{tab:detection-cross}
\small
\renewcommand{\arraystretch}{1.18}
\setlength{\tabcolsep}{2pt}
\begin{tabularx}{\textwidth}{@{}l*{3}{Y Y C{1.58cm}}@{}}
\toprule
\rowcolor{HeaderGray}
& \multicolumn{3}{c}{\textbf{Feature Squeezing}}
& \multicolumn{3}{c}{\textbf{Mahalanobis-Max}}
& \multicolumn{3}{c}{\textbf{ML-ATD}}\\
\cmidrule(lr){2-4}\cmidrule(lr){5-7}\cmidrule(l){8-10}
\rowcolor{HeaderGray}\textbf{Attack}
& AUROC & AUPR & TPR@5FPR
& AUROC & AUPR & TPR@5FPR
& AUROC & AUPR & TPR@5FPR\\
\midrule
\rowcolor{fatarow}\textbf{FATA} & 0.560 & 0.382 & 0.024 & 0.695 & 0.509 & 0.134 & 0.828 & 0.724 & 0.497\\
CAGE & 0.797 & 0.655 & 0.147 & 0.936 & 0.882 & 0.647 & 0.913 & 0.854 & 0.694\\
CAA & 0.564 & 0.385 & 0.027 & 0.833 & 0.765 & 0.468 & 0.802 & 0.721 & 0.533\\
Base & 0.665 & 0.494 & 0.064 & 0.813 & 0.660 & 0.278 & 0.952 & 0.918 & 0.801\\
\bottomrule
\end{tabularx}
\end{table}

%% file: 8-conclution.tex
\section{Conclusion}

FATA combines attention disruption and semantic preservation to shift damage
from full-token to compressed inference. It retains 96.3\% of LLaVA full-token
performance with 22.1\% practical-budget conditional blinding, and model-adapted
Qwen3.5 and InternVL3.5 tests show that the trigger extends beyond LLaVA. These
results call for evaluating token compression by adversarial behavior as well
as clean efficiency and accuracy.

\paragraph{Limitations.}
FATA assumes visual-encoder gradients. Its strongest Qwen3.5 and InternVL3.5
results use task- or architecture-aware extensions, with adaptation-dependent
full-token cost. Stealth is task-based, and ML-ATD is not an optimal defense.
Black-box transfer, broader tasks, and stronger online detectors remain open.

\section*{AI Use Statement}
Generative AI tools, including ChatGPT/Codex and Claude-based coding assistants,
were used to assist with implementation and debugging, dataset-processing
workflows, feedback on experimental procedures, and interpretation of
quantitative and qualitative results. They also supported literature search,
Chinese--English translation, manuscript drafting and editing, and preparation
of figures, tables, and references. AI-assisted revisions were checked against
the manuscript's equations and tables, dataset-selection records, and source
publications. The authors retain responsibility for the experimental evidence,
methodological choices, and final text, including all AI-assisted content.

\section*{Ethics Statement}
This work examines a security risk in efficient vision-language inference.
Attacks that preserve full-token behavior could be misused to evade screening
and disrupt systems after compression is enabled. Our evaluation uses existing
image-question benchmarks and model checkpoints to characterize this risk,
and includes detection experiments to inform mitigation.Deployment safety and robustness across
populations and applications require further evaluation. Reuse of benchmark
images and model checkpoints remains subject to their original licenses and privacy obligations; the attack
methods are intended for controlled, authorized robustness evaluation.

\section*{Reproducibility Statement}
Algorithm~\ref{alg:fata} and Appendix~\ref{app:method-details} specify the FATA
optimization procedure. Appendix~\ref{app:token-budgets} documents practical-budget selection and the
model-specific retention ratios. Appendix~\ref{app:metrics} defines the evaluation
metrics. Appendix~\ref{app:implementation} records the benchmark configuration
and attack settings, with the unique-image selection, deduplication, and
scoring protocol in Appendix~\ref{app:dataset-selection}.
The Qwen3.5 and InternVL3.5 adaptations are
given in Appendix~\ref{app:crossmodel}, and detector fitting, scoring, and
evaluation protocols are described in Appendix~\ref{app:detection-details}.

%% file: 9-appendix.tex

\section*{Appendix Contents}

\begingroup
\hypersetup{linkcolor=black}
\setlength{\parindent}{0pt}
\setlength{\parskip}{2pt}
\newcommand{\AppendixContentsLine}[2]{%
  \noindent
  \hyperref[#1]{%
    \makebox[\linewidth][l]{%
      \textbf{\underline{Appendix~\textcolor{red}{\ref*{#1}}}}.~%
      #2\nobreak\dotfill\nobreak\pageref*{#1}%
    }%
  }\par
}
\AppendixContentsLine{app:method-details}{Optimization Details}
\AppendixContentsLine{app:token-budgets}{Model-Specific Visual-Token Budgets}
\AppendixContentsLine{app:metrics}{Detailed Evaluation Definitions}
\AppendixContentsLine{app:implementation}{Experimental and Implementation Details}
\AppendixContentsLine{app:qualitative-cases}{Qualitative Compression-Triggered Cases}
\AppendixContentsLine{app:ablation}{Additional Ablation Results}
\AppendixContentsLine{app:crossmodel}{Additional Cross-Model Results}
\AppendixContentsLine{app:detection-details}{Detection Details and Cross-Attack Analysis}
\endgroup

\input{appendix/optimization_details}
\FloatBarrier
\input{appendix/token_budget_protocol}

\input{appendix/metrics_and_method_details}
\input{appendix/implementation_notes}
\clearpage
\input{appendix/qualitative_cases}
\clearpage
\input{appendix/additional_ablation}
\input{appendix/crossmodel_details}
\input{appendix/detection_details}

%% file: appendix/optimization_details.tex
\section{Optimization Details}
\label{app:method-details}

\subsection{Detailed Optimization Procedure}
Algorithm~\ref{alg:fata-detailed} presents the detailed optimization
procedure corresponding to Algorithm~\ref{alg:fata}. The target set $\mathcal M$ and its clean feature references are computed once and held fixed throughout optimization. At each iteration, attention scores and patch features are recomputed from the perturbed image to evaluate both losses on $\mathcal M$. Gradients are propagated through the frozen encoder to update
only the input perturbation.

\begin{algorithm}[ht]
\caption{FATA optimization in detail}
\label{alg:fata-detailed}
\small
\begin{algorithmic}[1]
\Require Clean image $x$, visual encoder $E$, budget $\epsilon$, step size $\alpha$, iterations $T$, weight $\lambda$, target count $M$
\State Extract clean attention $\mathbf s^{c}$ and patch features $\mathbf H^c$ from $E(x)$
\State Fix $\mathcal M\gets\TopK(\mathbf s^c,M)$
\State Initialize $\delta^{(0)}\sim\mathcal U(-\epsilon,\epsilon)$
\For{$t=0,\ldots,T-1$}
  \State $x_a^t\gets\Clip(x+\delta^{(t)},0,1)$
  \State Extract $\mathbf s^a,\mathbf H^a\gets E(x_a^t)$
  \State $\Lattn\gets\sum_{i\in\mathcal M}s_i^a$
  \State $\Lsem\gets1-\frac{1}{|\mathcal M|}\sum_{i\in\mathcal M}\cos(\mathbf h_i^a,\mathbf h_i^c)$
  \State $\Ltotal\gets\Lattn+\lambda\Lsem$
  \State $g^{(t)}\gets\nabla_\delta\Ltotal$
  \State $\delta^{(t+1)}\gets\Proj(\delta^{(t)}-\alpha\operatorname{sign}(g^{(t)}))$
\EndFor
\State \Return $\Clip(x+\delta^{(T)},0,1)$
\end{algorithmic}
\end{algorithm}

\FloatBarrier
\Needspace{0.55\textheight}
\subsection{Sensitivity to PGD Hyperparameters}
\label{app:hyperparameter-sensitivity}
Both hyperparameter sweeps evaluate FATA on LLaVA-1.5-7B using TextVQA-Open and VisionZIP.

\paragraph{Evaluation data.}
Tables~\ref{tab:alpha-sensitivity} and~\ref{tab:lambda} report experiments on unfiltered TextVQA-Open data used during method development, before applying the subset-selection procedure in Appendix~\ref{app:dataset-selection}. Table~\ref{tab:main-results} uses the filtered evaluation subset, which retains examples answered correctly with the original image but incorrectly with a black image at full tokens. This subset emphasizes reliance on visual evidence and exhibits more pronounced compression-triggered attack effects. The different sample populations therefore explain the difference in absolute accuracy between these sweeps and the TextVQA--VisionZIP results in Table~\ref{tab:main-results}, even when the model, compressor, and hyperparameters coincide.

We use random-start $\ell_\infty$ PGD with $\epsilon=2/255$, $T=100$, and $M=64$ fixed clean target tokens, and evaluate $K\in\{576,192,128,64,32,16\}$, with $K=576$ denoting full-token inference. We fix $\lambda=1$ when varying the step size $\alpha$ and $\alpha=0.5/255$ when varying the semantic-preservation weight $\lambda$. Step sizes are reported in units of $1/255$. Parameter selection considers full-token preservation and compressed accuracy over this trajectory.

\input{tables/table9_alpha_sensitivity}

\paragraph{PGD step size.}
At $\alpha=0.5/255$, Table~\ref{tab:alpha-sensitivity} reports 95.0\% full-token accuracy and the lowest mean compressed accuracy, 84.1\%, over $K\in\{192,128,64,32,16\}$. Reducing $\alpha$ to $0.25/255$ improves full-token accuracy to 96.0\% but raises the compressed-budget mean to 84.8\%. At $\alpha=0.75/255$, the corresponding values are 94.3\% and 84.4\%. We choose $\alpha=0.5/255$ for its stronger compression-triggered effect with high full-token accuracy. Its entries coincide with the shared $\lambda=1$ setting in Table~\ref{tab:lambda}.

\WFclear
\par\vspace{2pt}

\input{tables/table10_lambda_sensitivity}

\paragraph{Semantic-preservation weight.}
Table~\ref{tab:lambda} exposes the expected trade-off. Increasing $\lambda$ steadily improves full-token accuracy from 94.1\% at $\lambda=0.5$ to 96.0\% at $\lambda=8$, but it also raises the $K=16$ accuracy from 65.4\% to 70.7\%, indicating a weaker compression-triggered effect. The selected $\lambda=1$ recovers 0.9 points of full-token accuracy over $\lambda=0.5$ while retaining 68.8\% accuracy at $K=16$ and 80.1\% at $K=32$. Moving to $\lambda\geq2$ provides at most another 1.0 point at full tokens, but generally increases compressed-budget accuracy; at $\lambda=8$, for instance, the $K=32$ and $K=16$ accuracies rise to 81.0\% and 70.7\%. We choose $\lambda=1$ to limit full-token damage while retaining the compression-triggered effect that weakens at larger weights.

Together, the sweeps support $\alpha=0.5/255$ and $\lambda=1$ as a balance between full-token preservation and compression-triggered failure across budgets.

\WFclear
\FloatBarrier

%% file: tables/table9_alpha_sensitivity.tex
\begin{wraptable}[12]{r}{0.4\textwidth}
\vspace{-20pt}
\centering
\captionsetup{skip=3pt,justification=raggedright,singlelinecheck=false}
\caption{FATA accuracy on unfiltered TextVQA-Open across token budgets and step sizes $\alpha$.}
\label{tab:alpha-sensitivity}
\scriptsize
\setlength{\tabcolsep}{2.0pt}
\renewcommand{\arraystretch}{1.06}
\resizebox{\linewidth}{!}{%
\begin{tabular}{c|ccccccc}
\toprule
$K$ & .25 & .5 & .75 & 1 & 1.25 & 1.5 & 1.75\\
\midrule
576 & 96.0 & 95.0 & 94.3 & 94.5 & 94.2 & 95.1 & 94.3\\
192 & 94.1 & 92.7 & 93.4 & 93.8 & 93.8 & 92.9 & 94.0\\
128 & 93.0 & 91.5 & 92.1 & 92.1 & 92.8 & 91.7 & 92.7\\
64 & 88.9 & 87.6 & 88.4 & 88.5 & 89.4 & 88.8 & 88.7\\
32 & 78.8 & 80.1 & 80.7 & 80.6 & 82.0 & 81.4 & 80.9\\
16 & 69.4 & 68.8 & 67.3 & 69.4 & 68.1 & 69.2 & 71.0\\
\bottomrule
\end{tabular}%
}
\vspace{-7pt}
\end{wraptable}

%% file: tables/table10_lambda_sensitivity.tex
\begin{wraptable}[15]{r}{0.45\textwidth}
\vspace{-4pt}
\centering
\captionsetup{skip=3pt,justification=raggedright,singlelinecheck=false}
\caption{FATA accuracy on unfiltered TextVQA-Open across token budgets and semantic weights $\lambda$.}
\label{tab:lambda}
\small
\setlength{\tabcolsep}{3.2pt}
\renewcommand{\arraystretch}{1.06}
\resizebox{\linewidth}{!}{%
\begin{tabular}{c|ccccc}
\toprule
$K$ & $\lambda=.5$ & $1$ & $2$ & $4$ & $8$\\
\midrule
576 & 94.1 & 95.0 & 95.2 & 95.6 & 96.0\\
192 & 92.0 & 92.7 & 93.3 & 94.2 & 94.3\\
128 & 91.8 & 91.5 & 93.1 & 93.3 & 92.2\\
64 & 87.8 & 87.6 & 87.5 & 89.0 & 90.1\\
32 & 77.4 & 80.1 & 81.1 & 81.4 & 81.0\\
16 & 65.4 & 68.8 & 68.8 & 70.3 & 70.7\\
\bottomrule
\end{tabular}%
}
\vspace{-7pt}
\end{wraptable}

%% file: appendix/token_budget_protocol.tex
\section{Model-Specific Visual-Token Budgets}
\label{app:token-budgets}

\subsection{LLaVA-1.5-7B}
\label{app:llava-budgets}
LLaVA-1.5-7B receives a fixed sequence of 576 visual patch tokens from its vision encoder. We use $K=576$ for full-token inference and evaluate the five compressed budgets $K\in\{192,128,64,32,16\}$. Relative to the full sequence, these settings retain $1/3$, $2/9$, $1/9$, $1/18$, and $1/36$ of the visual tokens, respectively. Here, $K$ denotes the number of selected or merged visual representatives. In the reconstruction-based LLaVA path in Figure~\ref{fig:fata-overview}D, these representatives reconstruct an $N=576$-slot sequence before the projector. 

The practical budget is selected separately for each of the 16 dataset--compressor pairs. Let $\mathcal K=\{16,32,64,128,192\}$ denote the candidate budgets. Clean accuracy at each budget is measured relative to full-token clean accuracy:
\begin{equation}
R_{\mathrm{clean}}^{(d,c)}(K)=
\frac{\mathrm{Acc}_{\mathrm{Clean}}^{(d,c)}(K)}
{\mathrm{Acc}_{\mathrm{Clean}}^{(d,c)}(576)}.
\label{eq:clean-retention-ratio}
\end{equation}
Both accuracies are measured on the same examples using identical prompts and the same task scorer. Open-ended tasks use normalized VQA scores, and multiple-choice tasks use option-letter accuracy. Retention ratios are computed from unrounded task accuracies.

Using this ratio, we select the smallest candidate budget $K_{\mathrm{prac}}^{(d,c)}$ that retains at least 80\% of full-token clean accuracy:
\begin{equation}
K_{\mathrm{prac}}^{(d,c)}=
\min\left\{K\in\mathcal K:
R_{\mathrm{clean}}^{(d,c)}(K)\geq0.80\right\}.
\label{eq:llava-kprac-selection}
\end{equation}
The candidates are compared by their numeric values, from 16 upward. The uncompressed point $K=576$ is excluded from $\mathcal K$ because its retention ratio relative to itself is necessarily one.

Table~\ref{tab:llava-kprac} reports the practical budgets obtained by applying Eq.~\ref{eq:llava-kprac-selection} to each of the 16 dataset--compressor pairs.
\input{tables/table_llava_kprac}
The selected budgets are fixed for all attacks. All four compressors use $K=64$ on the open-ended tasks and $K=32$ on ScienceQA-MC. On VQAv2-MC, FlowCut requires $K=64$ for its clean accuracy at $K=32$ falls below 80\% of its full-token value; the other three compressors use $K=32$.

The visual-token retention ratio measures the fraction of visual tokens retained at the selected budget. For LLaVA, it is
\begin{equation}
r_{\mathrm{prac}}^{(d,c)}=
\frac{K_{\mathrm{prac}}^{(d,c)}}{576}.
\label{eq:llava-practical-ratio}
\end{equation}
$K=64$ and $K=32$ correspond to retention fractions of $1/9$ and $1/18$, respectively. CBR is evaluated at these budgets using the definition in Appendix~\ref{app:metrics}.

\subsection{Qwen3.5 and InternVL3.5}
\label{app:crossmodel-budgets}
The number of visual tokens in Qwen3.5 and InternVL3.5 varies across images because of their image preprocessing and token organization. For each image $i$ with $N_i$ visual tokens, the compression budget is specified as a retained fraction $r$ of $N_i$.

The main comparisons on Qwen3.5 and InternVL3.5 report full visual tokens, $1/3$ retention, and a practical budget specified for each task. The $1/3$ fraction matches the proportion retained by LLaVA at $K=192$. Before adversarial evaluation, the practical fractions are fixed at $1/9$ for TextVQA-Open and VQAv2-Open and $1/18$ for ScienceQA-MC and VQAv2-MC, corresponding to LLaVA's $64/576$ and $32/576$ settings, respectively. These fractions apply to all four compressors, with the visual token budgets and sample counts for all three models summarized in Table~\ref{tab:metric-inputs}.

\begin{table}[H]
\centering
\caption{Model-specific budgets and sample counts used in evaluation. Practical budgets are given separately for open-ended and multiple-choice tasks.}
\label{tab:metric-inputs}
\small
\setlength{\tabcolsep}{4pt}
\renewcommand{\arraystretch}{1.15}
\begin{tabularx}{\textwidth}{@{}l c Y Y l@{}}
\toprule
\textbf{Model} & \textbf{Full} & \makecell{\textbf{Practical}\\\textbf{Open}} & \makecell{\textbf{Practical}\\\textbf{MC}} & \makecell[l]{\textbf{Valid images}\\\textbf{per dataset}} \\
\midrule
LLaVA-1.5-7B & $576$ & $64$ & $32$ or $64$ & $1{,}000$ \\
Qwen3.5-9B & $N_i$ & $r=1/9$ & $r=1/18$ & $1{,}000$ \\
InternVL3.5-8B-HF & $N_i$ & $r=1/9$ & $r=1/18$ & $998$ or $1{,}000$ \\
\bottomrule
\end{tabularx}
\end{table}

%% file: tables/table_llava_kprac.tex
\begin{table}[H]
    \centering
    \caption{Clean-only practical token budgets
    $K_{\mathrm{prac}}^{(d,c)}$ for LLaVA-1.5-7B. Each entry is the smallest
    $K\in\{16,32,64,128,192\}$ that retains at least 80\% of the corresponding
    full-token clean accuracy.}
    \label{tab:llava-kprac}
    \small
    \setlength{\tabcolsep}{7pt}
    \renewcommand{\arraystretch}{1.12}
    \begin{tabular}{lcccc}
        \toprule
        \textbf{Compressor} & \textbf{TextVQA-Open} & \textbf{VQAv2-Open} &
        \textbf{ScienceQA-MC} & \textbf{VQAv2-MC} \\
        \midrule
        VisionZIP  & 64 & 64 & 32 & 32 \\
        VisPruner  & 64 & 64 & 32 & 32 \\
        PruMerge   & 64 & 64 & 32 & 32 \\
        FlowCut    & 64 & 64 & 32 & \textbf{64} \\
        \bottomrule
    \end{tabular}
\end{table}

%% file: appendix/metrics_and_method_details.tex
\section{Detailed Evaluation Definitions}
\label{app:metrics}

The metrics in Section~\ref{sec:preliminaries} use the model-specific budgets in Table~\ref{tab:metric-inputs}. Accuracy enters the equations as a fraction in $[0,1]$ and is reported as a percentage.
The metrics in Section~\ref{sec:preliminaries} are evaluated using the budgets and sample counts summarized in Table~\ref{tab:metric-inputs}.

\paragraph{Task scores and binary correctness.}
For a dataset with $n$ valid examples, let $v_i\in[0,1]$ denote the task credit and $c_i\in\{0,1\}$ the binary correctness of prediction $i$. The corresponding accuracies are
\begin{equation}
\mathrm{Acc}^{\mathrm{task}}=\frac{1}{n}\sum_{i=1}^{n}v_i,
\qquad
\mathrm{Acc}^{\mathrm{EM}}=\frac{1}{n}\sum_{i=1}^{n}c_i.
\label{eq:task-and-binary-accuracy}
\end{equation}
Table~\ref{tab:task-scoring-rules} gives the task-specific rules. Open answers use the benchmark's normalization of punctuation, articles, and number words before matching.

\begin{table}[H]
\centering
\caption{Task-level scoring rules for reported accuracy and binary correctness. CBR uses the binary decisions in the final column.}
\label{tab:task-scoring-rules}
\small
\setlength{\tabcolsep}{5pt}
\renewcommand{\arraystretch}{1.15}
\begin{tabularx}{\textwidth}{@{}p{2.5cm} X X@{}}
\toprule
\textbf{Task} & \textbf{Task credit $v_i$} & \textbf{Binary correctness $c_i$} \\
\midrule
TextVQA-Open and VQAv2-Open & Normalized VQA credit against the human reference answers, with partial credit allowed. & $1$ for a normalized exact match to an accepted reference answer; $0$ otherwise. \\
\addlinespace
ScienceQA-MC & $1$ if the parsed letter (A--F) matches the annotated answer index converted to a letter; $0$ otherwise. & The same option-letter decision as $v_i$. \\
\addlinespace
VQAv2-MC & $1$ if the parsed letter (A--D) matches the fixed target derived from the modal human answer; $0$ otherwise. & The same option-letter decision as $v_i$. \\
\bottomrule
\end{tabularx}
\end{table}


\paragraph{Metric computation.}
The accuracy tables for all three models report $\mathrm{Acc}^{\mathrm{task}}$. SR is computed using this accuracy for LLaVA and Qwen3.5 and $\mathrm{Acc}^{\mathrm{EM}}$ for InternVL3.5, and is averaged over the four datasets. 

CER is reported only for LLaVA and is the complement of $\mathrm{Acc}^{\mathrm{task}}$ at $K_{\mathrm{prac}}$, following Eq.~\ref{eq:cer}. It retains partial task credit and measures overall error without conditioning on Clean correctness or isolating attack-induced failures. The summary is the equal-weight mean over the 16 dataset--compressor settings, calculated from the displayed accuracies, expressed as a percentage, and rounded to two decimal places, as specified alongside Table~\ref{tab:stealth-lethality}.

CBR uses the binary correctness decisions $c_i$ for all three models, following Eq.~\ref{eq:cbr}. For attack $m$, its denominator requires $c_i=1$ for Clean at full tokens, attack $m$ at full tokens, and Clean under the evaluated compressor and practical budget; the numerator additionally requires $c_i=0$ for attack $m$ under that same compression setting. Macro CBR is the unweighted mean over the 16 dataset--compressor pairs, whereas pooled CBR is computed by summing their numerator counts and denominator counts before division.

%% file: appendix/implementation_notes.tex
\section{Experimental and Implementation Details}
\label{app:implementation}

\paragraph{Main benchmark.}
The main victim is LLaVA-1.5-7B with a CLIP ViT-Large visual encoder. We test VisionZIP, VisPruner, PruMerge, and FlowCut at $K\in\{576,192,128,64,32,16\}$. Each of TextVQA-Open, VQAv2-Open, ScienceQA-MC, and VQAv2-MC contains 1,000 evaluation samples. Base and FATA use $\epsilon=2/255$, $\alpha=0.5/255$, and 100 steps; CAA uses the same $\epsilon=2/255$ comparison budget with $\alpha=1/255$ and 100 steps; CAGE is run under $\epsilon=2/255$. 

\FloatBarrier
\subsection{Evaluation Subset Construction}
\label{app:dataset-selection}

The procedure below documents the released dataset reconstruction. The reported experiments retain their archived evaluation mappings; in particular, the original VQAv2-MC options cannot be reconstructed from the released builder's seed alone.

\paragraph{Selection unit and representative question.}
Each subset uses unique images as the selection unit. Images are deduplicated using source identifiers and image content. For each image, the first eligible question in source order is fixed before model inference. If the resulting pair fails the selection criteria, no alternative question is selected for that image.

\paragraph{Selection based on model responses.}
LLaVA decodes each candidate greedily with at most 32 new tokens, using the same prompt and question for the original image and a $336\times336$ black image. Both conditions use all 576 visual tokens and greedy decoding with at most 32 new tokens. We retain an image only if the answer is incorrect with a $336\times336$ black image and correct with the original image. Let $q_i$ denote the representative question assigned to image $x_i$, $b$ the black image, and $c(\hat y,y)$ the correctness rule used during subset construction. The evaluation subset is

\begin{equation}
\mathcal D_j^{\mathrm{eval}}
=\operatorname{First}_{1000}\!\left\{(x_i,q_i,y_i)\in\mathcal U_j:
c(f_{576}(b,q_i),y_i)=0\ \land\
c(f_{576}(x_i,q_i),y_i)=1\right\},
\label{eq:dataset-filter}
\end{equation}

where $\mathcal U_j$ contains the candidate question--image pairs and reference answers for dataset $j$, ordered by source occurrence after image deduplication. $\operatorname{First}_{1000}$ selects the first 1,000 pairs satisfying both conditions. Selection uses no attack predictions or results under token compression.

\paragraph{TextVQA-Open.}
We use the TextVQA validation split. During selection, open answers are lowercased, punctuation and articles are removed, and number words from zero through five are mapped to digits. A prediction passes if it exactly matches a normalized reference answer or contains that answer within the generated phrase.

\paragraph{VQAv2-Open.}
We use the VQAv2 validation split and exclude yes/no questions. Selection uses the same answer normalization and matching rule as TextVQA-Open.

\paragraph{VQAv2-MC.}
The released reconstruction uses the VQAv2 validation split, excludes yes/no questions and invalid option sets, and pairs the modal human answer with three distinct distractors. Its configured global seed is $s_0=20260904$. For each question, the UTF-8 string formed by \texttt{vqav2-mc-options-v1}, $s_0$ in decimal, and \texttt{question\_id}, separated by null bytes, is hashed with SHA-256. The first eight digest bytes, interpreted as an unsigned big-endian integer, seed a local Python \texttt{random.Random} instance for distractor draws and option shuffling. Thus $s_0$ is a fixed configuration constant, while each question's random seed is derived deterministically from it and the question ID. The reconstruction excludes images selected for VQAv2-Open. For the archived experiments, the original option-generation seed and RNG state are unavailable; exact reproduction uses the saved questions, option order, and target letters in the archived mapping.

\paragraph{ScienceQA-MC.}
The candidate pool combines the ScienceQA train, validation, and test splits for evaluation only. We discard records without images or with invalid metadata or choices. Image deduplication also removes exact duplicates across splits. The optional hint is prepended to the question, the original choices are labeled A--F, and the annotated answer index determines the target letter.

\paragraph{Relation to reported accuracy.}
The reported evaluations use archived question--image mappings with their saved answer specifications; the released reconstruction above does not redefine those samples. Evaluation uses each model's prompt and formal task scorer, so clean full-token accuracy can be below 100\%. Task scoring and metric computation are detailed in Appendix~\ref{app:metrics}.

\subsection{Performance across Token Budgets}

\begin{figure}[ht]
    \centering
    \includegraphics[width=\linewidth]{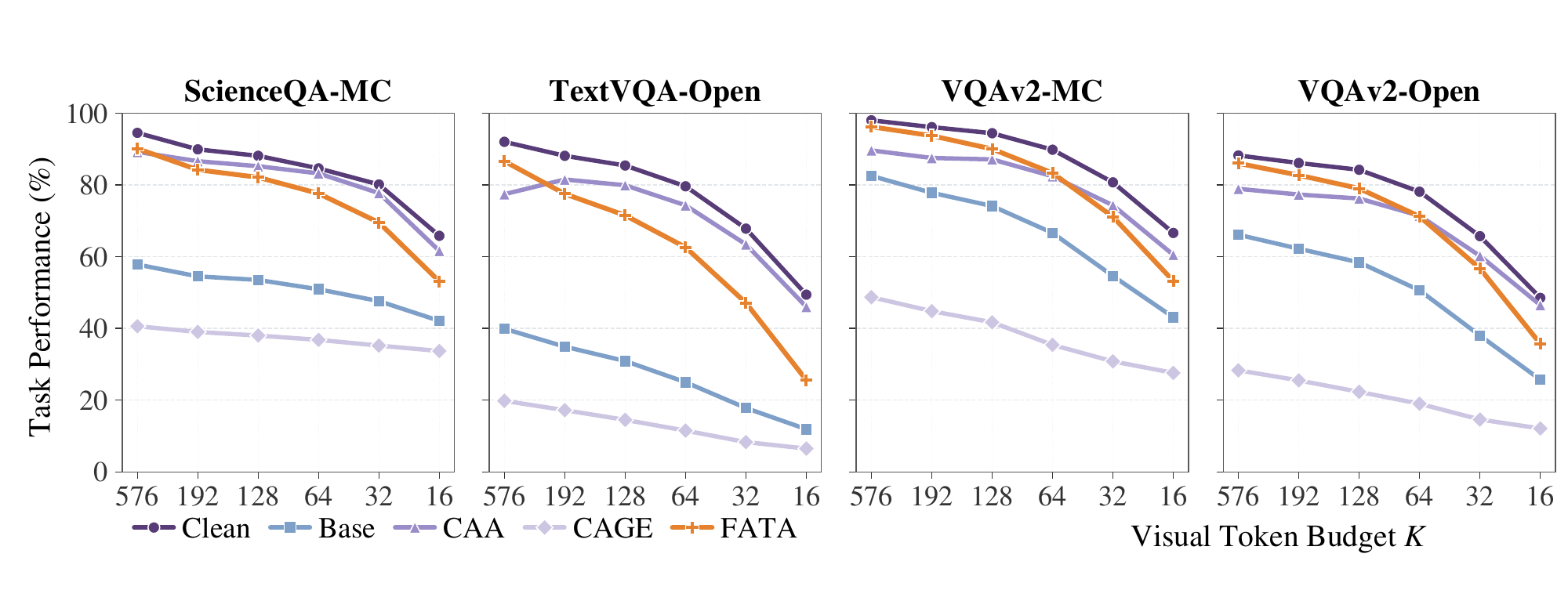}
    \caption{Average task performance versus visual-token budget. The exact
    per-method values are reported in Table~\ref{tab:main-results} in the main paper.}
    \label{fig:budget-curves}
\end{figure}

\FloatBarrier
Figure~\ref{fig:budget-curves} visualizes the results in Table~\ref{tab:main-results} of the main paper for each dataset. The point at $K=576$ reports full-token accuracy, and each compressed-budget point averages accuracy over VisionZIP, VisPruner, PruMerge, and FlowCut.

TextVQA-Open shows the largest Clean--FATA gap at every compressed budget. Under severe compression, both open-ended tasks reach lower Clean and FATA accuracies than the multiple-choice tasks. Across all four datasets, the Clean--FATA gap widens as the token budget decreases.

\FloatBarrier

%% file: appendix/qualitative_cases.tex
\section{Qualitative Compression-Triggered Cases}
\label{app:qualitative-cases}
In the LLaVA examples, selection maps show last-layer CLIP CLS-to-patch attention on a shared monotonic log scale that preserves token rankings. This visual-encoder attention is independent of the language input. Each Clean Top-$K$ panel partitions all $K$ clean tokens into those present in the adversarial Top-$K$ set (gold) and those displaced (red).

\subsection{LLaVA-1.5-7B}
\label{app:cases-llava}
\noindent\begin{minipage}{\textwidth}
\centering
\includegraphics[width=\textwidth]{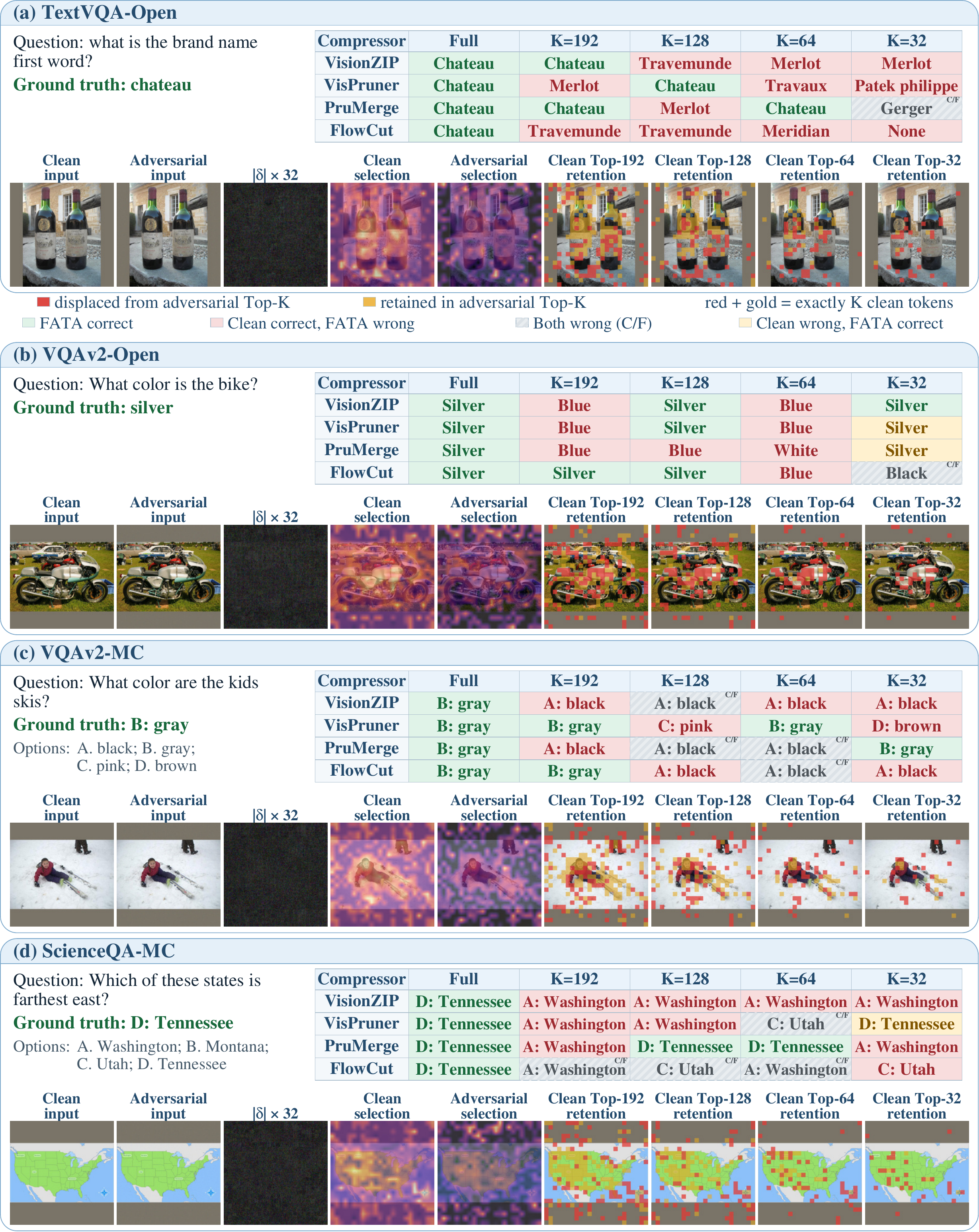}
\captionsetup{hypcap=false,skip=4pt}
\captionof{figure}{LLaVA-1.5-7B cases from TextVQA-Open (a), VQAv2-Open (b), VQAv2-MC (c), and ScienceQA-MC (d). Tables report FATA predictions across compressors and token budgets, with the shared legend below (a) applying to all panels.}
\label{fig:case-llava}
\end{minipage}\par

\clearpage
\subsection{Qwen3.5-9B}
\label{app:cases-qwen35}
Selection maps visualize the $\ell_2$ norm of post-merger visual features on a shared logarithmic scale. The retention panels track the corresponding global Top-$K$ sets; each compressor follows its own selection rule. Full-token predictions remain correct for Clean and FATA in all four cases. Compressed responses depend on both the compressor and retention fraction. In VQAv2-Open, VisionZIP recovers the correct FATA answer at retention $1/9$ after errors at the two larger fractions.

\noindent\begin{minipage}{\textwidth}
\centering
\includegraphics[width=\textwidth]{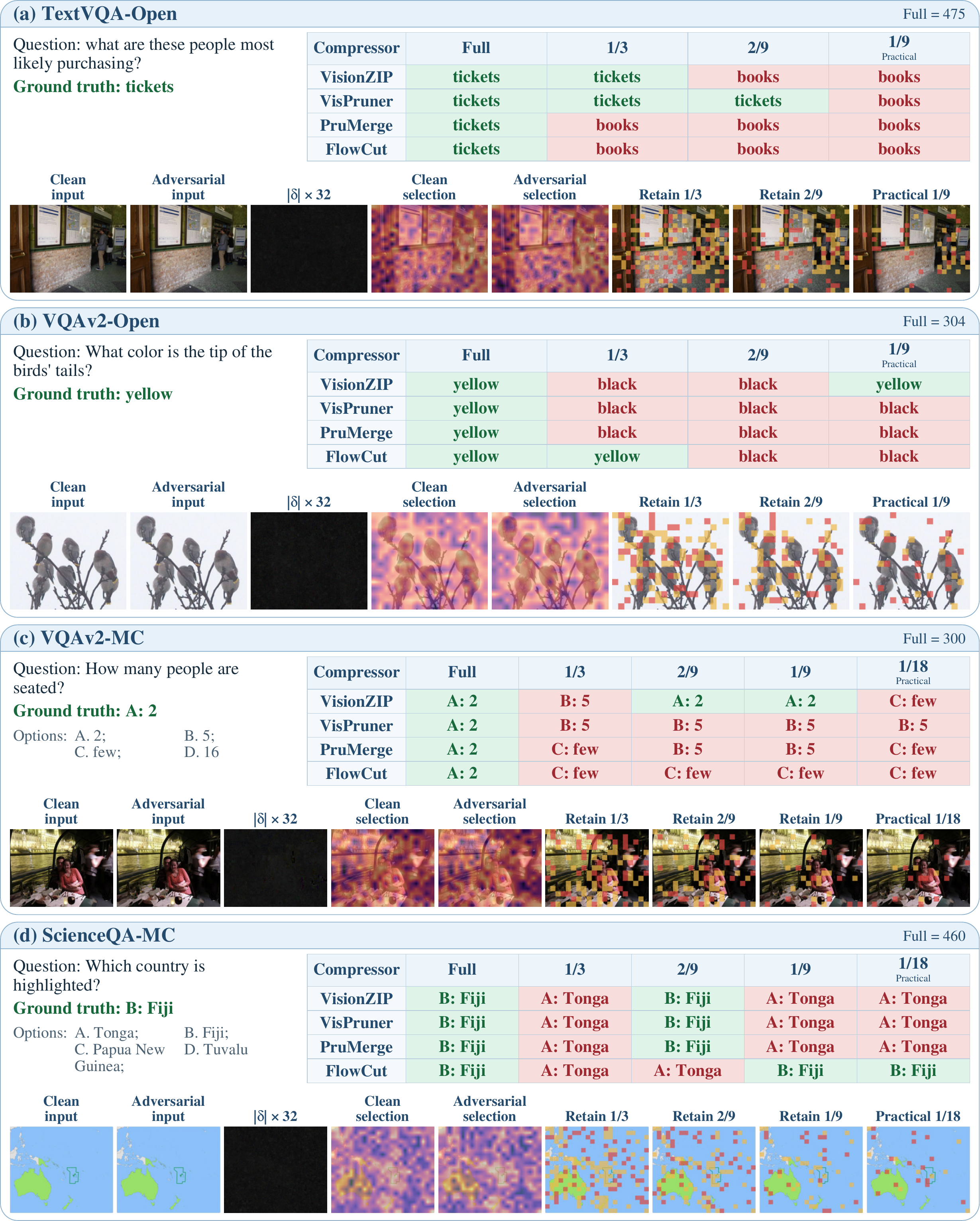}
\captionsetup{hypcap=false,skip=4pt}
\captionof{figure}{Qwen3.5-9B cases in the same task order as Figure~\ref{fig:case-llava}. Tables report FATA answers using the shared colour conventions.}
\label{fig:case-qwen35}
\end{minipage}\par

\clearpage
\subsection{InternVL3.5-8B-HF}
\label{app:cases-internvl35}
InternVL processes local image tiles and a thumbnail, whose CLS-to-patch scores and token footprints are projected onto the original image. In VQAv2-MC, clean support for the shirt comes mainly from thumbnail tokens. At practical retention, all four compressors lose their target-overlapping clean anchors and return incorrect FATA answers.

\noindent\begin{minipage}{\textwidth}
\centering
\includegraphics[width=\textwidth,height=0.84\textheight,keepaspectratio]{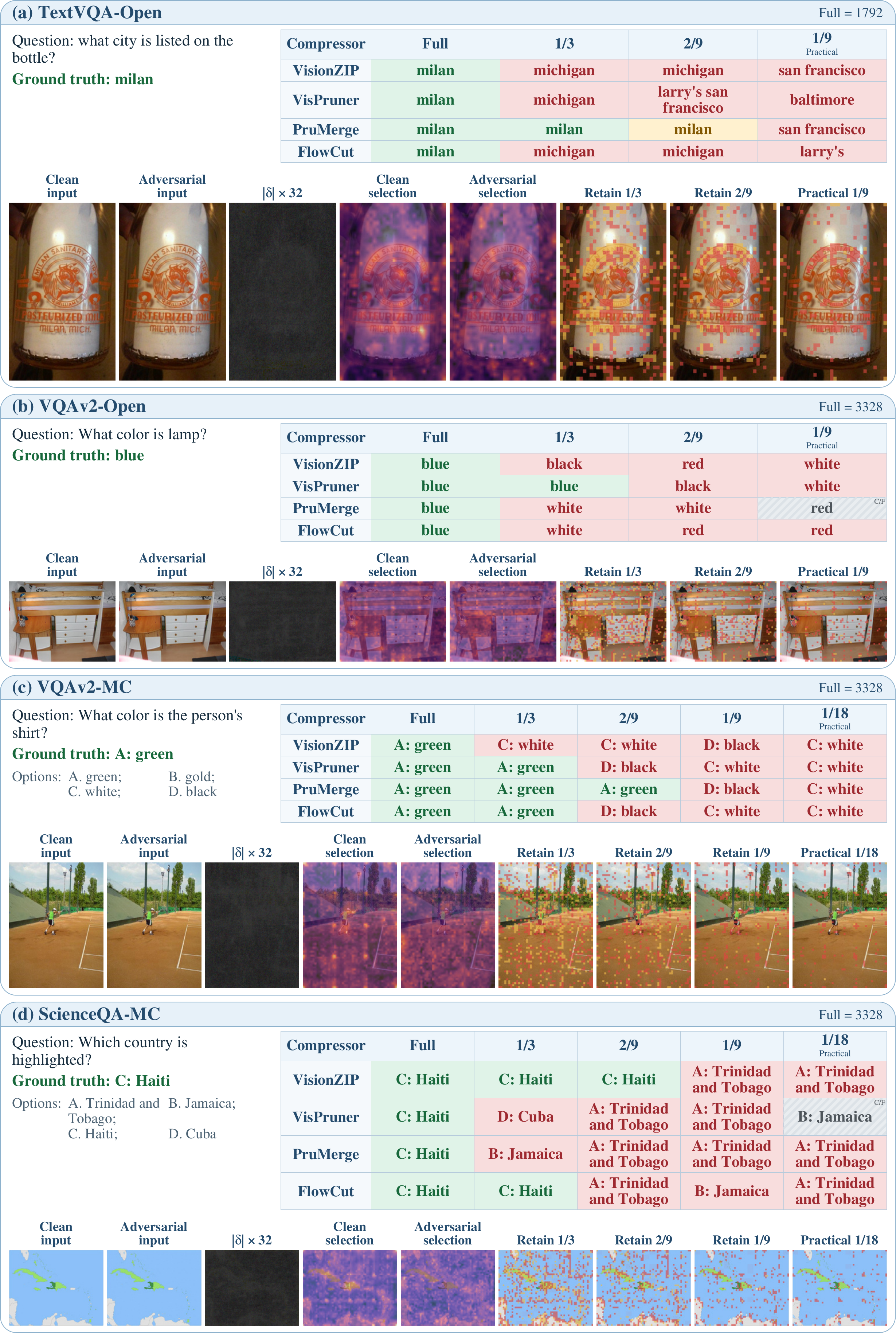}
\captionsetup{hypcap=false,skip=4pt}
\captionof{figure}{InternVL3.5-8B-HF cases across the four tasks, with global native-priority Top-$K$ panels and the colour conventions of Figure~\ref{fig:case-llava}.}
\label{fig:case-internvl35}
\end{minipage}\par

%% file: appendix/additional_ablation.tex
\section{Additional Ablation Results}
\label{app:ablation}
\Needspace{0.38\textheight}
\begin{wrapfigure}[16]{r}{0.49\textwidth}
    \vspace{-4pt}
    \centering
    \includegraphics[width=0.96\linewidth]{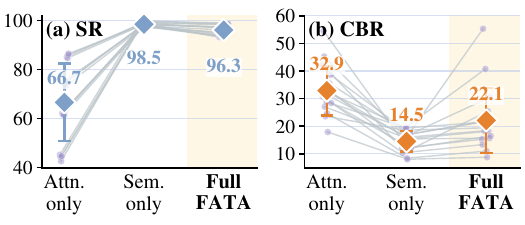}
    \captionsetup{hypcap=false,justification=raggedright,singlelinecheck=false}
    \caption{Loss ablation across 16 dataset--compressor settings, with paired lines and mean diamonds; shading marks full FATA. Error bars show $\pm1$ sample standard deviation across settings}
    \label{fig:ablation-global}
    \vspace{-8pt}
\end{wrapfigure}

Figure~\ref{fig:ablation-global} shows how the loss variants behave within individual dataset and compressor settings. In panel (a), attention-only SR spans roughly 40--90\%, whereas semantic-only values cluster near 100\%. Full FATA keeps all plotted SR values above 90\%, showing that semantic preservation also reduces variation across settings. Panel (b) shows a range for full FATA CBR, from roughly 9\% to 55\%. Most settings remain below 30\%, with two near 40\% and 55\%. The trajectories from semantic-only to full FATA include both increases and decreases in CBR, so the aggregate gain does not hold uniformly across settings. Together, the panels show consistent full-token retention under the joint objective and a more variable compression-triggered effect.

\WFclear

%% file: appendix/crossmodel_details.tex
\section{Additional Cross-Model Results}
\label{app:crossmodel}

This appendix details the model-specific objectives and supplementary analyses for the cross-model experiments in Section~\ref{sec:cross-model}. Visual-token budgets and scoring rules are specified in Appendices~\ref{app:token-budgets} and~\ref{app:metrics}, respectively.

\subsection{Qwen3.5 task-aware FATA adaptation}
\label{app:qwen-adaptation}
The Qwen3.5 adaptation uses task-level losses to preserve full-token predictions while inducing errors under compression, following FATA's design principle. This white-box adaptation uses ground-truth answers and downstream Qwen3.5 gradients.

\paragraph{Task-aware objective.}
Let $x_a$ be the perturbed input, $y$ the ground-truth answer, $m_t$ the compressor selected at PGD step $t$, and $r_t$ the retained-token fraction used at that step. In the released Qwen implementation, $x$ denotes the processor's pixel-value tensor and $x_a=\operatorname{clip}(x+\delta,-1,1)$. The four compressors are cycled round-robin, and the budget schedule emphasizes the practical operating point while revisiting $1/3$ retention. We write the compression-side attack score as
\begin{equation}
\mathcal A_t =
 w_{\rm task}\,\mathcal L_{\rm task}^{m_t,r_t}
 + w_{\rm rank}\,\mathcal D_{\rm rank}
 + w_{\rm comp}\,\mathcal D_{\rm comp},
\label{eq:qwen-attack-branch}
\end{equation}
where $\mathcal L_{\rm task}^{m_t,r_t}$ is the teacher-forced ground-truth cross-entropy under compressed inference, written as a score to be maximized. The locked coefficients are $w_{\rm task}=1.5$, $w_{\rm rank}=0.25$, and $w_{\rm comp}=0.05$.

\paragraph{Rank and compressor auxiliaries.}
The $N$ merged tokens in \texttt{pooler\_output} are denoted by $z_i(x)$. Rank the clean tokens by descending $\|z_i(x)\|_2$. For each of the medium and practical budgets, $K=\max(1,\operatorname{round}(Nr))$ with $r=1/3$ and $r=1/9$ (Open) or $1/18$ (MC), respectively, take
\begin{equation}
w_K=\min\!\left\{\max\!\left(8,\operatorname{round}(0.05N)\right),K,N-K\right\}
\label{eq:qwen-boundary-window}
\end{equation}
tokens on each side of the cutoff: the core set contains clean ranks $K-w_K+1,\ldots,K$, and the decoy set contains ranks $K+1,\ldots,K+w_K$. Let $C$ and $D$ be the respective deduplicated unions over the two budgets. Both sets are fixed before optimization and reused at every step.

\Needspace{7\baselineskip}
Let $\overline v_S=|S|^{-1}\sum_{i\in S}v_i$ and $n_i=\|z_i(x_a)\|_2$, and let $\mathcal M$ contain VisionZIP, VisPruner, FlowCut, and PruMerge. The released implementation defines the maximization scores as
\begin{align}
\mathcal D_{\rm rank}
&=-\operatorname{softplus}\!\left(0.05+\overline n_C-\overline n_D\right),
\label{eq:qwen-rank-definition}\\
\mathcal D_{\rm comp}
&=-\frac{1}{4}\sum_{m\in\mathcal M}
\operatorname{softplus}\!\left(0.05+\overline s_{m,C}-\overline s_{m,D}\right).
\label{eq:qwen-comp-definition}
\end{align}
Thus, both margins are $0.05$, and softplus is applied after taking the core and decoy means. The negative signs convert the implemented boundary losses into scores to maximize. Unlike the task term, these auxiliaries use the fixed two-budget targets; the compressor term averages all four surrogates at every step.

For completeness, the surrogate token scores are
\begin{align}
s_{\rm VisionZIP}&=0.75a+0.25v,&
s_{\rm VisPruner}&=0.5a+0.5b,\notag\\
s_{{\rm FlowCut},i}&=(a_i+c_i)\|z_i\|_1,&
s_{\rm PruMerge}&=0.7a+0.3p.
\label{eq:qwen-surrogate-scores}
\end{align}
Here all features are evaluated at $x_a$. Define $\operatorname{Norm}_1(u)=u/(\sum_i u_i+10^{-8})$, $a=\operatorname{Norm}_1(n)$, and $b=\operatorname{Norm}_1(d)$, where $d_i=1-(\sum_j\cos(z_i,z_j)-1)/(N-1)$. The distinctiveness score is $v_i=1-\max_{j\in T_{\rm dom}}\cos(z_i,z_j)$, where $T_{\rm dom}$ contains the $\max(1,\lfloor N/4\rfloor)$ current tokens with largest feature norms. With $\bar z=N^{-1}\sum_jz_j$, $c=\operatorname{Norm}_1(([\cos(z_i,\bar z)]_+)_i)$. Finally, $p=b$ for $N\le1024$ and $p=\operatorname{Norm}_1((1/N,\ldots,1/N))$ otherwise. The PruMerge surrogate receives no spatial grid in this call, and all four surrogate temperatures equal one.

Full-token preservation is imposed with two complementary terms. The first keeps the ground-truth answer likely at full tokens,
\begin{equation}
\mathcal L_{\rm full}=\operatorname{CE}\bigl(p_{\theta}(\cdot\mid x_a,\mathrm{Full}),y\bigr),
\label{eq:qwen-full-ce}
\end{equation}
and the second distills the clean full-token predictive distribution,
\begin{equation}
\mathcal L_{\rm distill}=D_{\rm KL}
\left(p_{\theta}(\cdot\mid x,\mathrm{Full})\;\|\;
      p_{\theta}(\cdot\mid x_a,\mathrm{Full})\right).
\label{eq:qwen-distill}
\end{equation}
Both are computed with teacher forcing during optimization; autoregressive generation is reserved for evaluation. The preservation branch is
\begin{equation}
\mathcal P = w_{\rm full}\mathcal L_{\rm full}
           + w_{\rm distill}\mathcal L_{\rm distill},
\qquad
w_{\rm full}=w_{\rm distill}=1.
\label{eq:qwen-preservation}
\end{equation}
The two branches summarize the ascent and preservation signs as
\begin{equation}
\mathcal J_t(\delta)=\mathcal A_t-\mathcal P.
\label{eq:qwen-total-objective}
\end{equation}
The actual update separately normalizes each term's gradient before weighting, rather than differentiating this fixed weighted sum. With $\mathcal N(g)=g/(\|g\|_2+10^{-8})$, using the L2 norm over the entire gradient tensor, its direction is
\begin{align}
g_t={}&1.5\,\mathcal N(\nabla_\delta\mathcal L_{\rm task}^{m_t,r_t})
+0.25\,\mathcal N(\nabla_\delta\mathcal D_{\rm rank})
+0.05\,\mathcal N(\nabla_\delta\mathcal D_{\rm comp})\notag\\
&-\mathcal N(\nabla_\delta\mathcal L_{\rm full})
-\mathcal N(\nabla_\delta\mathcal L_{\rm distill}).
\label{eq:qwen-gradient-combination}
\end{align}
Each gradient includes the derivative mask of the input clipping operation. Random-start $\ell_\infty$ PGD then updates
\begin{equation}
\delta^{(t+1)}=
\Pi_{[-\epsilon,\epsilon]}
\left[\delta^{(t)}+\alpha\,\operatorname{sign}
\left(g_t\right)\right],
\qquad
x_a^{(t+1)}=\operatorname{clip}(x+\delta^{(t+1)},-1,1),
\label{eq:qwen-pgd}
\end{equation}
with $\epsilon=2/255$, $\alpha=0.5/255$, and $T=100$ steps in the processor input domain. The final attack uses no additional gradient-projection operator. One adversarial image is generated per input and shared across all four compressors and evaluation budgets.

The compression branch aims to alter which visual evidence survives compression. The preservation branch uses full-token cross-entropy and clean-logit distillation to preserve task semantics under Qwen3.5's dynamic tokenization.

\paragraph{Dataset-level amplification.}
Table~\ref{tab:qwen-dataset-amplification} complements the aggregate results in Table~\ref{tab:qwen} with a dataset-level breakdown. Here $D_r$ denotes the Clean--FATA accuracy difference at budget $r$, and $A_r=D_r-D_{\rm Full}$. Both are calculated before rounding. Practical-budget amplification is positive across all four datasets.

\begin{table}[htbp]
\centering
\caption{Qwen3.5 compression-specific damage and amplification by dataset
(percentage points). Positive amplification means that compression increases
the attack-induced accuracy gap relative to full-token inference.}
\label{tab:qwen-dataset-amplification}
\small
\setlength{\tabcolsep}{4.5pt}
\begin{tabular}{lrrrrr}
\toprule
\textbf{Dataset} & $D_{\rm Full}$ & $D_{1/3}$ & $D_{\rm Prac}$ & $A_{1/3}$ & $A_{\rm Prac}$ \\
\midrule
TextVQA-Open & 0.70 & 5.22 & 8.60 & 4.52 & 7.90 \\
VQAv2-Open & 1.00 & 10.00 & 14.57 & 9.00 & 13.57 \\
ScienceQA-MC & -0.20 & 9.47 & 23.70 & 9.67 & 23.90 \\
VQAv2-MC & 1.00 & 8.05 & 19.60 & 7.05 & 18.60 \\
\bottomrule
\end{tabular}
\end{table}


\subsection{InternVL3.5 architecture-aware FATA adaptation}
\label{app:internvl-adaptation}
The InternVL3.5 adaptation combines compression-side disruption with preservation of the uncompressed representation. Its archived V5 objective includes VisionZIP, VisPruner, DivPrune, PruMerge, and FlowCut; the reported comparisons show four of these compressors, excluding DivPrune.

\paragraph{Features and post-compression disruption.}
For an image with $J$ tiles, let $H(x)\in\mathbb R^{J\times256\times d_h}$ contain the post-shuffle features before the projector, and let $P(x)\in\mathbb R^{J\times256\times d_p}$ contain the projector outputs. Thus $N=256J$; all tiles, including a thumbnail when present, contribute patch tokens, with CLS excluded. For compressor $c$ and retention fraction $r$, the attack pass produces $Z_{c,r}(x)$ with $K_{c,r}$ output tokens. The post-compression score compares corresponding output positions:
\begin{equation}
\mathcal D_{\rm pcd}^{c,r}
=1-\frac{1}{K_{c,r}}\sum_{j=1}^{K_{c,r}}
\cos\!\left(z_j^{c,r}(x_a),z_j^{c,r}(x)\right).
\label{eq:internvl-pcd}
\end{equation}
The cosine is computed along the feature dimension and then averaged over tokens, without pooling beforehand. This attack pass disables reconstruction; PruMerge returns selected centers without merging in this mode. The comparison follows output order rather than matching source-token identities. Clean outputs are fixed, while adversarial selections are recomputed at each step.

\paragraph{Cutoff targets and retained sets.}
At the practical fraction $r_*$, the total budget is $K=\max(1,\operatorname{round}(Nr_*))$. Each tile first receives $\lfloor K/J\rfloor$ tokens, with the remainder assigned to earlier tiles. For a tile budget $k_t$, sort the clean family scores $s_{c,t,i}(x)$ in descending order and set
\begin{equation}
b_t=\min\!\left\{64,\max\!\left(16,\operatorname{round}(0.25k_t)\right),k_t-1,255-k_t\right\}.
\label{eq:internvl-boundary-window}
\end{equation}
The ordered lists $\mathcal B^+_{c,t}$ and $\mathcal B^-_{c,t}$ contain ranks $k_t-b_t+1,\ldots,k_t$ and $k_t+1,\ldots,k_t+b_t$, respectively. Each has $b_t$ entries: $16$ for Open and $13$ or $14$ for MC at the practical fractions. They are fixed from the clean image throughout optimization, without combining different budgets. Ties use the descending \texttt{argsort} order, with no extra tie rule. Separately, $\mathcal U_{c,t}$ contains the source-token indices actually selected by clean compressor $c$ in tile $t$ at $r_*$. Its size is the actual number of selected representatives, with no additional cap. These retained sets are also fixed and need not equal the boundary lists or their union.

The ranking score pairs entries at the same position in the two lists:
\begin{equation}
\mathcal D_{\rm rank}^{c}
=-\frac1J\sum_{t=1}^{J}\frac1{b_t}\sum_{j=1}^{b_t}
\left[s_{c,t,\mathcal B^+_{c,t}[j]}(x_a)
      -s_{c,t,\mathcal B^-_{c,t}[j]}(x_a)\right]_+,
\qquad [u]_+=\max(u,0).
\label{eq:internvl-rank}
\end{equation}
The margin is $\mu=0$: this is a paired ReLU penalty expressed as a score to maximize. All reported tiles have nonempty boundary lists; an empty list is omitted from the tile mean.

\paragraph{Family scores.}
Write $A$ for the number of attention heads and $\alpha_{a,i}$ for penultimate-layer CLS attention after summing each four-patch group into token $i$. With $u_i=A^{-1}\sum_a\alpha_{a,i}$, the VisionZIP and VisPruner rank scores are $\sum_a\alpha_{a,i}$ and $u_i$, respectively. PruMerge uses final-layer CLS query--key attention, while FlowCut combines normalized attention, value similarity, and value magnitude:
\begin{align}
s_{{\rm PruMerge},i}
&=\operatorname{softmax}_{j=0,\ldots,N}
\left(q_0^\top k_j/\sqrt{d_k}\right)_i,\notag\\
e_i&=\frac1A\sum_a\operatorname{softmax}_{j=0,\ldots,N}
\left(v_{a,0}^\top v_{a,j}\right)_i,\notag\\
s_{{\rm FlowCut},i}
&=\left[\operatorname{Norm}(u)_i+\operatorname{Norm}(e)_i\right]
\left\|\frac1A\sum_a v_{a,i}\right\|_1,
\label{eq:internvl-family-scores}
\end{align}
where $\operatorname{Norm}(w)=w/(\sum_{i=1}^{N}w_i+10^{-8})$. In these two archived rank helpers, index $0$ is the first tile's CLS and indices $1,\ldots,N$ span the image-wide patch stream; actual compressor passes remain tile-local. The Q/K/V features are aligned by concatenating the four patch-group slots and repeating CLS to the matching width $d_k$ or $d_v$. FlowCut uses penultimate-layer values, with no scale or temperature in its value softmax; CLS is dropped before $\operatorname{Norm}$. DivPrune uses $s_{{\rm DivPrune},t,i}=\min_{j\in\mathcal U_{{\rm DivPrune},t}}[1-\cos(h_{t,i},h_{t,j})]$, allowing self-matches. Evaluating these scores on clean features fixes the cutoff lists; their adversarial values enter Eq.~\ref{eq:internvl-rank}.

\paragraph{Diversity and relation/value terms.}
For $c\in\{\mathrm{DivPrune},\mathrm{VisPruner}\}$, the diversity score increases nearest-neighbor similarity within the fixed retained set:
\begin{equation}
\mathcal D_{\rm div}^{c}
=\frac1J\sum_t\frac1{|\mathcal U_{c,t}|}
\sum_{i\in\mathcal U_{c,t}}
\max_{j\in\mathcal U_{c,t}\setminus\{i\}}
\hat h_{t,i}(x_a)^\top\hat h_{t,j}(x_a),
\quad \hat h=\frac{h}{\max(\|h\|_2,10^{-12})}.
\label{eq:internvl-div}
\end{equation}
Self-pairs are excluded, and tiles with fewer than two selected tokens are omitted from the mean. The pairwise cosine distance is $1-\hat h_i^\top\hat h_j$; Eq.~\ref{eq:internvl-div} averages each selected token's largest non-self similarity.

The relation term uses FlowCut's selected-token value magnitude. Let $V_{a,i}(x_a)$ be the penultimate-layer value of global patch token $i$ in head $a$, define $\nu_i=\|A^{-1}\sum_a V_{a,i}(x_a)\|_1$, and map the tile-local retained sets into their image-wide union $\mathcal U_{\rm FlowCut}$. Then
\begin{equation}
\mathcal D_{\rm rel}
=-\frac{|\mathcal U_{\rm FlowCut}|^{-1}
\sum_{i\in\mathcal U_{\rm FlowCut}}\nu_i}
{\max\!\left(N^{-1}\sum_{i=1}^{N}\nu_i,10^{-6}\right)}.
\label{eq:internvl-rel}
\end{equation}
Head averaging precedes the feature L1 norm. Normalization divides the selected-token mean by the mean over all adversarial patch tokens, with a denominator floor of $10^{-6}$. Gradients propagate through both means.

\paragraph{Attention and redundancy targets.}
Let $a_{t,i}$ average CLS attention over heads and the last three vision layers, then average each four-patch group into a post-shuffle token. In each tile, the clean ranking fixes the top $q=\operatorname{round}(256\cdot64/576)=28$ targets $\mathcal T_t$, the next $28$ decoys $\mathcal E_t$, and width-$8$ lists $\mathcal A^\pm_{\rho,t}$ on either side of rank $\operatorname{round}(256\rho)$ for $\rho\in\mathcal R=\{1/3,1/9,1/18\}$. Let $\langle\cdot\rangle$ average over tiles and corresponding list positions. The attention score is
\begin{equation}
\begin{split}
\mathcal D_{\rm attn}={}&-\langle a_{\mathcal T}(x_a)\rangle
-\langle[a_{\mathcal T}(x_a)-a_{\mathcal E}(x_a)]_+\rangle\\
&-\frac13\sum_{\rho\in\mathcal R}
\langle[a_{\mathcal A^+_\rho}(x_a)-a_{\mathcal A^-_\rho}(x_a)]_+\rangle.
\end{split}
\label{eq:internvl-attn}
\end{equation}
The redundancy score compares the same targets with the other $228$ projected tokens within their tile:
\begin{equation}
\mathcal D_{\rm red}
=\frac1{Jq}\sum_t\sum_{i\in\mathcal T_t}
\max_{j\notin\mathcal T_t}\cos\!\left(p_{t,i}(x_a),p_{t,j}(x_a)\right).
\label{eq:internvl-red}
\end{equation}
These target/complement sets differ from the compressor-specific retained sets $\mathcal U_{c,t}$.

\Needspace{10\baselineskip}
\paragraph{Semantic preservation and pooling.}
The semantic hierarchy consists of target tokens, tile means, and an image mean of the projector output. Pooling is arithmetic averaging: $\operatorname{pool}(\{p_i\}_{i\in S})=|S|^{-1}\sum_{i\in S}p_i$. Define $\bar p_t=256^{-1}\sum_i p_{t,i}$ and $\bar p=N^{-1}\sum_{t,i}p_{t,i}$. The three levels receive equal weight:
\begin{equation}
\begin{split}
\mathcal L_{\rm sem}=\frac13\Bigg[&
\frac1{Jq}\sum_t\sum_{i\in\mathcal T_t}
\left(1-\cos(p_{t,i}(x_a),p_{t,i}(x))\right)\\
&+\frac1J\sum_t\left(1-\cos(\bar p_t(x_a),\bar p_t(x))\right)
+1-\cos(\bar p(x_a),\bar p(x))\Bigg].
\end{split}
\label{eq:internvl-sem}
\end{equation}
Tile and image means include all full-token patch positions and any thumbnail tile, without CLS. Equal tile sizes make the mean of tile means identical to the mean over all $N$ tokens. PCD, semantic, and projected-redundancy cosines use a feature-norm floor of $10^{-8}$. Pooling applies to the two semantic means; PCD compares compressed tokens individually.

\paragraph{Complete objective and optimization.}
For Open tasks, the nonzero budget weights are $\omega_d(1/9)=1$, $\omega_d(1/3)=0.4$, and $\omega_d(2/9)=0.25$. For MC tasks they are $\omega_d(1/18)=1$, $\omega_d(1/9)=0.4$, and $\omega_d(1/3)=0.2$. Only PCD uses the normalized multibudget combination:
\begin{equation}
\mathcal D_{\rm pcd}
=\frac{\sum_r\omega_d(r)\,|\mathcal C_r|^{-1}
\sum_{c\in\mathcal C_r}\mathcal D_{\rm pcd}^{c,r}}
{\sum_r\omega_d(r)}.
\label{eq:internvl-pcd-aggregation}
\end{equation}
Here $\mathcal C_{r_*}$ contains all five attack compressors; secondary budgets exclude DivPrune. Rank averages the five practical-budget scores, diversity averages DivPrune and VisPruner, and relation uses FlowCut. The attention and redundancy terms use their fixed targets above. The maximization form is
\begin{align}
\max_{\|\delta\|_\infty\le\epsilon}\quad
\mathcal J_{\rm InternVL}
={}&6\,\mathcal D_{\rm pcd}
+3\,(20\,\mathcal D_{\rm rank})
+2\,\mathcal D_{\rm div}+2\,\mathcal D_{\rm rel}\notag\\
&+40\,\mathcal D_{\rm attn}
+1.5\,\mathcal D_{\rm red}-0.05\,\mathcal L_{\rm sem}.
\label{eq:internvl-total}
\end{align}
These scores reverse the minimized implementation losses, up to additive constants for PCD, diversity, and redundancy. The single weighted objective is differentiated directly, without separately normalizing or projecting term gradients. Random-start PGD uses
\begin{equation}
\delta^{(t+1)}=\Pi_{[-\epsilon,\epsilon]}
\left[\delta^{(t)}+\alpha\,\operatorname{sign}
\left(\nabla_\delta\mathcal J_{\rm InternVL}\right)\right],
\qquad x_a^{(t+1)}=\operatorname{clip}(x+\delta^{(t+1)},0,1),
\label{eq:internvl-pgd}
\end{equation}
with $\epsilon=2/255$, $\alpha=0.5/255$, and $100$ steps. The random seed for zero-based sample index $i$ is $42+i$.

\paragraph{Additional retention levels.}
Table~\ref{tab:internvl-full-trajectory} supplements Table~\ref{tab:internvl} with additional retention fractions applied uniformly across all four datasets. Damage is the Clean--FATA accuracy difference, and amplification is its increase relative to full-token inference. Both are calculated before rounding.

\begin{table}[htbp]
\centering
\caption{InternVL3.5 accuracy and amplification at additional retention fractions. Each row applies the same fraction to all four datasets and averages over the four compressors.}
\label{tab:internvl-full-trajectory}
\small
\setlength{\tabcolsep}{6pt}
\begin{tabular}{lrrrr}
\toprule
\textbf{Budget} & \textbf{Clean} & \textbf{FATA} & \textbf{Damage (pp)} & \textbf{Amplification (pp)} \\
\midrule
$2/9$ & 80.77 & 68.49 & 12.28 & 7.05 \\
$1/9$ & 70.78 & 56.20 & \textbf{14.58} & \textbf{9.34} \\
$1/18$ & 60.07 & 47.74 & 12.33 & 7.09 \\
$1/36$ & 49.97 & 41.90 & 8.06 & 2.83 \\
\bottomrule
\end{tabular}
\end{table}

The Clean--FATA accuracy gap peaks at $1/9$ and narrows as retention falls to $1/36$.

\FloatBarrier
\Needspace{17\baselineskip}
\paragraph{Retained-set diagnostics.}
For either diagnostic $m$ and a fixed compressor $c$, macro averaging first gives equal weight to valid images within each dataset, then equal weight to the four datasets:
\begin{equation}
\bar m_c=\frac14\sum_{d=1}^{4}
\left(\frac1{n_d}\sum_{i\in\mathcal I_d}m_{i,d,c}\right).
\label{eq:internvl-diagnostic-macro}
\end{equation}
Here $\mathcal I_d$ is the valid-image set and $n_d=|\mathcal I_d|$. In the order TextVQA-Open, VQAv2-Open, ScienceQA-MC, and VQAv2-MC, the counts are $998,1000,1000,1000$. No weighting by token count is used. For Table~\ref{tab:internvl-compressor-diagnostics}, tile-local selected indices are mapped into the concatenated image-wide token stream before computing each image's diagnostics. Both diagnostics use the final adversarial input at the practical budget ($1/9$ for Open and $1/18$ for MC).

\begin{table}[H]
\centering
\caption{InternVL3.5 retained-set diagnostics at the practical budget. For each image, $S_{\rm clean}$ and $S_{\rm FATA}$ contain the selected source-token indices, using representative indices for merging methods. Flip rate is $|S_{\rm clean}\setminus S_{\rm FATA}|/|S_{\rm clean}|$, the fraction of clean-selected tokens removed after attack; Jaccard is $|S_{\rm clean}\cap S_{\rm FATA}|/|S_{\rm clean}\cup S_{\rm FATA}|$. Values are fractions, macro-averaged by Eq.~\ref{eq:internvl-diagnostic-macro}.}
\label{tab:internvl-compressor-diagnostics}
\small
\setlength{\tabcolsep}{6pt}
\begin{tabular}{lrr}
\toprule
\textbf{Compressor} & \textbf{Flip rate} & \textbf{Jaccard} \\
\midrule
VisionZIP & 0.854 & 0.079 \\
VisPruner & 0.925 & 0.040 \\
FlowCut & 0.890 & 0.059 \\
PruMerge & 0.617 & 0.240 \\
\bottomrule
\end{tabular}
\end{table}

The retained-set statistics are consistent with substantial changes to the compression decision across the retained compressors, while the family-aware surrogates cover complementary attention-, diversity-, and relation-sensitive selection behavior. These diagnostics do not establish that any single surrogate is solely responsible for the task-level drop.

\paragraph{InternVL3.5 data integrity.}
The TextVQA-Open exclusions summarized in Appendix~\ref{app:token-budgets} arise from truncated source images 371 and 759, which are omitted consistently from Clean and FATA aggregation. The final formal validation reports no duplicate or missing conditions, no NaN or Inf values, and no perturbation-budget violations. The small set of parser status exceptions is inherited from the same fixed evaluation pipeline and is scored as zero credit without re-running inference.

%% file: appendix/detection_details.tex
\section{Detection Details and Cross-Attack Analysis}
\label{app:detection-details}

\subsection{ML-ATD Construction}
\begin{figure}[ht]
    \centering
    \includegraphics[width=\textwidth]{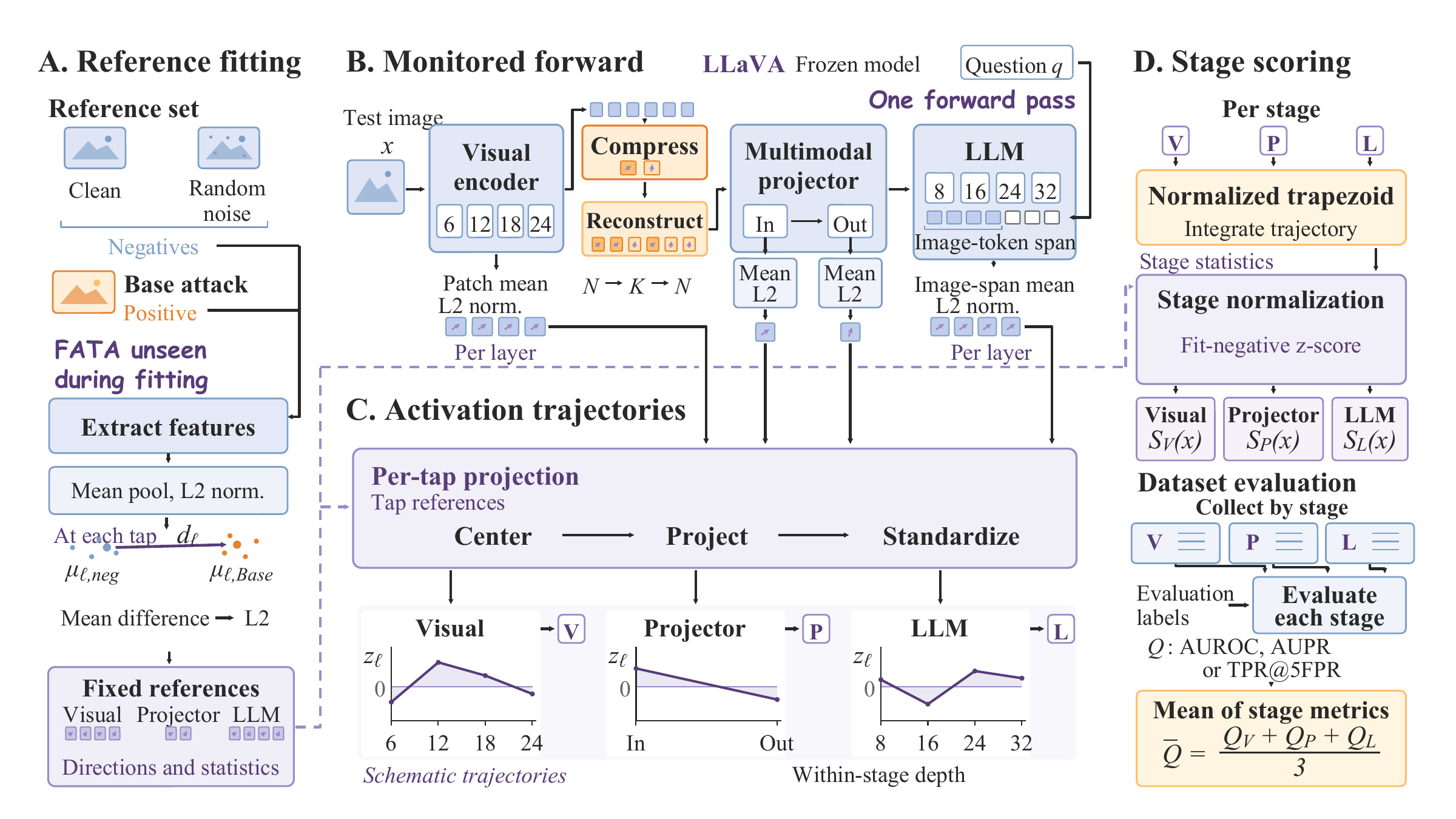}
    \caption{ML-ATD reference fitting, feature readout, and stage-wise detection.
    Visual, Projector, and LLM trajectories are scored and evaluated separately,
    with stage metrics averaged only for reporting.}
    \label{fig:mlatd}
\end{figure}

HiddenDetect~\citep{jiang2025hiddendetect} characterizes jailbreak behavior through directions derived from refusal-related hidden-state changes. Inspired by HiddenDetect, ML-ATD adapts this approach by replacing the refusal-related reference with an attack direction estimated from Base examples and extending feature readout to the visual encoder, projector, and language model. Figure~\ref{fig:mlatd} shows reference fitting, feature extraction, trajectories within each stage, and the calculation of detection metrics. Each stage is evaluated in its own feature space.

\paragraph{Fixed per-tap references (A).}
Reference and test images use the same feature extraction procedure, with mean pooling followed by $\ell_2$ normalization at each observation point, or tap, $\ell$, giving $\widetilde{\mathbf f}_\ell(x)$. The negative references, denoted by \texttt{clean\_clip+random\_clip}, consist of clean images and images generated by adding random perturbations to clean images, without filtering by prediction correctness. Base adversarial examples provide the positive references. The mean feature vectors of the positive and negative references at each tap define the fixed direction

\begin{equation}
    \mathbf d_\ell=
    \frac{\boldsymbol\mu_{\ell,\mathrm{Base}}-
    \boldsymbol\mu_{\ell,\mathrm{neg}}}
    {\|\boldsymbol\mu_{\ell,\mathrm{Base}}-
    \boldsymbol\mu_{\ell,\mathrm{neg}}\|_2}.
\end{equation}
Reference fitting uses only the negative references and Base examples, excluding FATA examples.

\paragraph{Monitored feature readout (B).}
A single forward pass through the frozen LLaVA model provides visual patch features from layers 6, 12, 18, and 24, the projector input and output, and LLM activations from layers 8, 16, 24, and 32. Mean pooling covers all patches for visual features and only the image-token span for LLM features. The projector input and output each use mean pooling and $\ell_2$ normalization, with separate reference directions and calibration statistics. In the compressed inference path shown in Figure~\ref{fig:mlatd}, token selection or merging produces $K$ representatives $Z$, which reconstruct an $N$-slot sequence $\widetilde H$ before the projector. The LLM receives the projected visual sequence together with the question.

\paragraph{Standardized projections (C).}
Each normalized feature is centered on its negative mean, projected onto its own attack direction, and standardized
\begin{equation}
    u_\ell(x)=\bigl(\widetilde{\mathbf f}_\ell(x)-
    \boldsymbol\mu_{\ell,\mathrm{neg}}\bigr)^\top\mathbf d_\ell,
    \qquad
    z_\ell(x)=\frac{u_\ell(x)-m_\ell^-}{s_\ell^-},
\end{equation}
where $m_\ell^-$ and $s_\ell^-$ denote the mean and standard deviation of $u_\ell$ over the negative reference samples. For each stage $r\in\{V,P,L\}$, $\mathbf z_r(x)$ contains the standardized projections in layer order, with depth normalized to $[0,1]$ within that stage. Schematic in panel C illustrates how these values vary across layers.

\paragraph{Stage scores and reported metrics (D).}
ML-ATD integrates the standardized projections over the normalized depth interval using the trapezoidal rule. The resulting area $T_r(x)$ gives the detection score for stage $r$ after normalization by its mean $a_r^-$ and standard deviation $b_r^-$ over the negative reference samples,
\begin{equation}
    T_r(x)=\operatorname{Trapz}_{[0,1]}\!\bigl(\mathbf z_r(x)\bigr),
    \qquad
    S_r(x)=\frac{T_r(x)-a_r^-}{b_r^-}.
\end{equation}

Evaluation uses the scores $S_V(x)$, $S_P(x)$, and $S_L(x)$ separately to compute AUROC, AUPR, and TPR@5FPR against the attack and negative labels. For each stage, TPR@5FPR is obtained from its test-set ROC curve at a 5\% false positive rate. For each metric $Q$, the reported value is the mean across the three stages,
\begin{equation}
    \overline Q=\frac{Q_V+Q_P+Q_L}{3}.
\end{equation}

\Needspace{14\baselineskip}
\noindent\textbf{FATA--Base comparison.}\par
\vspace{2pt}
\input{tables/table8_detection_main}

Table~\ref{tab:detection-main} isolates the FATA and Base rows of the main-text comparison in Table~\ref{tab:detection-cross}; the green arrows report relative reductions from Base to FATA. At a $5\%$ false positive rate, FATA reduces TPR by $62.5\%$ under Feature Squeezing (from $6.4\%$ to $2.4\%$), by $51.8\%$ under Mahalanobis-Max (from $27.8\%$ to $13.4\%$), and by $37.9\%$ under ML-ATD (from $80.1\%$ to $49.7\%$, with the reduction computed before rounding). Thus, FATA is less detectable than Base across all three detectors at this operating point.

\subsection{Detection Evaluation}

\paragraph{Baseline settings and aggregation.}
The results in Tables~\ref{tab:detection-cross} and~\ref{tab:detection-main} use two negatives per positive attack example. Feature Squeezing (FS) evaluates inputs at $K=576$ using the unthresholded MaxFS score (\texttt{max\_fs\_score\_diff}), which measures changes in task credit against ground-truth answers before and after input transformations. Mahalanobis-Max uses the per-image maximum of standardized CLS and mean-pooled patch distances in CLIP space, with 128-dimensional PCA, clean fitting at indices 0--499, and evaluation at 500--999. Both baselines use seed 0, without attack-specific score selection.

FS averages 16 dataset--compressor combinations with identical full-token scores across compressors, making this equivalent to averaging four datasets. Mahalanobis-Max combines matching records across compressors before averaging four datasets. ML-ATD reports an equal-weight mean over four compressors (VisionZIP, VisPruner, PruMerge, and FlowCut), four datasets, and the three stages. Its detection runs use $K=64$ for open-ended tasks and $K=32$ for multiple-choice tasks.

\paragraph{Input consistency versus feature deviation.}
The four-attack comparison in Table~\ref{tab:detection-cross} shows that detection performance depends on the type of deviation each detector measures. In particular, Mahalanobis-Max orders CAGE $>$ CAA $>$ Base $>$ FATA on all three metrics. Although FS gives similar detection performance for FATA and CAA, Mahalanobis-Max detects FATA less readily. This lower detectability in feature space is consistent with FATA's semantic preservation objective.

\paragraph{Evaluation sample coverage.}
CAA evaluation on ScienceQA includes 990 attack examples for FS and 496 for Mahalanobis-Max, with 1,980 and 992 aligned negatives, respectively. Other conditions use nominal counts.

\clearpage

%% file: tables/table8_detection_main.tex
\begin{center}
\let\uparrow\downarrow
\captionsetup{hypcap=false}
\captionof{table}{Detection performance on FATA and Base. Higher values mean easier detection; green arrows show the relative reduction from Base to FATA.}
\label{tab:detection-main}
\small
\renewcommand{\arraystretch}{1.10}
\setlength{\tabcolsep}{7.5pt}
\begin{tabularx}{\textwidth}{l l >{\centering\arraybackslash}X >{\centering\arraybackslash}X >{\centering\arraybackslash}X}
\toprule
\rowcolor{HeaderGray}\textbf{Detector} & \textbf{Attack} & \textbf{AUROC} & \textbf{AUPR} & \textbf{TPR@5FPR}\\
\midrule
Feature Squeezing & FATA & \textbf{0.560}\,\stealthgain{15.8} & \textbf{0.382}\,\stealthgain{22.7} & \textbf{0.024}\,\stealthgain{62.5}\\
& Base & 0.665 & 0.494 & 0.064\\
\midrule
Mahalanobis-Max & FATA & \textbf{0.695}\,\stealthgain{14.5} & \textbf{0.509}\,\stealthgain{22.9} & \textbf{0.134}\,\stealthgain{51.8}\\
& Base & 0.813 & 0.660 & 0.278\\
\midrule
\rowcolor{fatarow}ML-ATD & FATA & \textbf{0.828}\,\stealthgain{13.0} & \textbf{0.724}\,\stealthgain{21.1} & \textbf{0.497}\,\stealthgain{37.9}\\
\rowcolor{fatarow}& Base & 0.952 & 0.918 & 0.801\\
\bottomrule
\end{tabularx}
\end{center}